\documentclass{article}

\usepackage{microtype}
\usepackage{graphicx}
\usepackage{subfigure}
\usepackage{booktabs} 

\usepackage{hyperref}

\usepackage[accepted]{icml2020}

\newtheorem{problem}{Problem}

\newtheorem{theorem}{Theorem}

\usepackage{amsmath,amsfonts}
\usepackage{amssymb} 
\usepackage{dsfont} 
\usepackage{enumitem}
\usepackage{pgfplots}
\usepackage{nicefrac}

\newcommand{\myblue}[1]{#1}

\icmltitlerunning{Born-Again Tree Ensembles}

\begin{document}

\twocolumn[
\icmltitle{Born-Again Tree Ensembles \\}



\icmlsetsymbol{equal}{*}

\begin{icmlauthorlist}
\icmlauthor{Thibaut Vidal}{puc}
\icmlauthor{Toni Pacheco}{puc}
\icmlauthor{Maximilian Schiffer}{tum}
\end{icmlauthorlist}

\icmlaffiliation{puc}{Department of Computer Science, Pontifical Catholic University of Rio de Janeiro (PUC-Rio), Rio de Janeiro, Brazil.}
\icmlaffiliation{tum}{TUM School of Management, Technical University of Munich, Munich, Germany}

\icmlcorrespondingauthor{Thibaut Vidal}{vidalt@inf.puc-rio.br}

\icmlkeywords{Random Forests, Interpretability, Simplification, Dynamic Programming}

\vskip 0.3in
]



\printAffiliationsAndNotice{}  
\begin{abstract}
The use of machine learning algorithms in finance, medicine, and criminal justice can deeply impact human lives. As a consequence, research into interpretable machine learning has rapidly grown in an attempt to better control and fix possible sources of mistakes and biases. Tree ensembles offer a good prediction quality in various domains, but the concurrent use of multiple trees reduces the interpretability of the ensemble. Against this background, we study born-again tree ensembles, i.e., the process of constructing a single decision tree of minimum size that reproduces the exact same behavior as a given tree ensemble \myblue{in its entire feature space}. To find such a tree, we develop a dynamic-programming based algorithm that exploits sophisticated pruning and bounding rules to reduce the number of recursive calls. This algorithm generates optimal born-again trees for many datasets of practical interest, leading to classifiers which are typically simpler and more interpretable without any other form of compromise.
\end{abstract}

\section{Introduction}
\label{sec:introduction}

Tree ensembles constitute a core technique for prediction and classification tasks. Random forests \citep{Breiman2001} and boosted trees \citep{Friedman2001} \myblue{have been} used in various application fields, e.g., in medicine for recurrence risk prediction and image classification, in criminal justice for custody decisions, or in finance for credit risk evaluation. Although tree ensembles offer a high prediction quality, distorted predictions in high-stakes decisions can be exceedingly harmful. Here, interpretable machine learning models are essential to understand potential distortions and biases. \myblue{Research} in this domain \myblue{has significantly increased} \citep{MurdochSinghEtAl2019} with numerous works focusing on the construction of optimal sparse trees \citep{HuRudinEtAl2019} or on the interpretability of neural networks \citep{ZhangNianWuEtAl2018,MelisJaakkola2018}.

Currently, there exists a trade-off between the interpretability and the performance of tree (ensemble) classifiers. \myblue{Single} decision trees (e.g., those produced by CART) are well-known for their interpretability, whereas tree ensembles and gradient boosting approaches allow for high prediction quality but are generally more opaque and redundant. Against this background, we study born-again tree ensembles in a similar notion as born-again trees \citep[see,][]{BreimanShang1996}, and search for a simpler classifier that faithfully reproduces the behavior of a tree ensemble.

Formally, let $(\mathbf{X},\mathbf{y}) = \{\mathbf{x}_i,y_i\}_{i=1}^n$ be a training set in which each $\mathbf{x}_i \in \mathbb{R}^p$ is a $p$-dimensional numerical feature vector, and each $y_i \in \mathbb{N}$ is its associated class. Each sample of this training set has been independently drawn from an unknown distribution $(\mathcal{X},\mathcal{Y})$. Based on this training set, a tree ensemble $\mathcal{T}$ learns a function $F_\mathcal{T}:\mathcal{X}\rightarrow \mathcal{Y}$ that predicts $y_i$ for each $\mathbf{x}_i$ drawn from $\mathcal{X}$. With \myblue{this notation}, we state Problem~\ref{prob:bornagain}, which is the core of our studies.

\begin{problem}[Born-again tree ensemble]
\label{prob:bornagain}
Given a tree ensemble $\mathcal{T}$, we search for a decision tree $T$ of \textbf{minimal size} \myblue{that is \textbf{faithful} to $\mathcal{T}$, i.e.,} such that $F_T(\mathbf{x}) = F_\mathcal{T}(\mathbf{x})$ for all $\mathbf{x} \in \mathbb{R}^p$.
\end{problem}

We note that the condition $F_T(\mathbf{x}) = F_\mathcal{T}(\mathbf{x})$ applies to the entire feature space. Indeed, our goal is to faithfully reproduce the \myblue{decision function} of the tree ensemble for all possible inputs in $\mathcal{X}$.
In other words, we are looking for a \emph{new representation of the same classifier}.
Problem~\ref{prob:bornagain} depends on the definition of a \emph{size} metric. In this study, we refer to the size of a tree either as its \emph{depth} \myblue{(D)} or \myblue{its} \emph{number of \myblue{leaves}} \myblue{(L)}. Additionally, we study a hierarchical objective \myblue{(DL) which optimizes depth in priority and then the number of leaves}.
For brevity, we detail the methodology for the \emph{depth} objective (D) in the main paper. The \myblue{supplementary material contains the algorithmic adaptations needed to cover the other objectives, rigorous proofs for all theorems, as well as additional illustrations and experimental results}.

Theorem~\ref{theorem:complexity} states the computational complexity of Problem~\ref{prob:bornagain}. \pagebreak
\begin{theorem}\label{theorem:complexity}
Problem~\ref{prob:bornagain} is \myblue{NP-hard} when optimizing depth, number of \myblue{leaves}, or any hierarchy of these two objectives.
\end{theorem}
This result uses a direct reduction from 3-SAT. \myblue{Actually, the same proof shows that the sole fact of verifying the faithfulness of a solution is NP-hard}. In this work, we show that despite this \myblue{intractability result}, Problem~\ref{prob:bornagain} can be solved \emph{to proven optimality} for various datasets of practical interest, and that the solution of this problem permits significant advances regarding tree ensemble \myblue{simplification}, interpretation, and \myblue{analysis}.

\subsection{State of the Art}
\label{subsec:stateoftheart}

Our work relates to the field of interpretable machine learning, especially thinning tree ensembles and optimal decision tree construction. We review these fields concisely and refer to \citet{GuidottiMonrealeEtAl2018}, \citet{MurdochSinghEtAl2019} and \citet{Rudin2019} for surveys and discussions on interpretable machine learning, as well as to \citet{Rokach2016} for an overview on general work on decision forests.

\paragraph{Thinning tree ensembles} \hspace*{-0.3cm}
has been studied from different perspectives and divides in two different streams, \textit{i}) classical thinning of a tree ensemble by removing some weak learners from the original ensemble and \textit{ii}) replacing a tree ensemble by a simpler classifier, e.g., a \myblue{single decision} tree. 

Early works on thinning focused on finding reduced ensembles which yield a prediction quality comparable to the full ensemble \citep{MargineantuDietterich1997}. Finding such reduced ensembles has been proven to be NP-hard \citep{TamonXiang2000} and in some cases reduced ensembles may even outperform the full ensemble \citep{ZhouWuEtAl2002}. While early works proposed a static thinning, dynamic thinning algorithms that store the full ensemble but dynamically query only a subset of the trees have been investigated by \citet{Hernandez-LobatoMartinez-MuozEtAl2009}, \citet{ParkFurnkranz2012}, and \citet{Martinez-MunozHernandez-LobatoEtAl2008}. For a detailed discussion on this stream of research we refer to \citet{Rokach2016}, who discusses the development of ranking-based methods \citep[see, e.g.,][]{ProdromidisStolfoEtAl1999,CaruanaNiculescu-MizilEtAl2004,BanfieldHallEtAl2005,HuYuEtAl2007,PartalasTsoumakasEtAl2010,Rokach2009,ZhangWang2009} and search-based methods \citep[see, e.g.,][]{ProdromidisStolfo2001,WindeattArdeshir2001,ZhouWuEtAl2002,ZhouTang2003,RokachMaimonEtAl2006,ZhangBurerEtAl2006}. 

In their seminal work about born-again trees, \citet{BreimanShang1996} were the first to introduce a thinning problem that aimed at replacing a tree ensemble by a newly constructed simpler classifier. Here, they used a tree ensemble to create a data set which is then used to \myblue{build} a born-again tree with a prediction accuracy close to the accuracy of the tree ensemble. Ensuing work followed three different concepts. \citet{Meinshausen2010} introduced the concept of \textit{node harvesting}, i.e., reducing the number of decision nodes to generate an interpretable tree. Recent works along this line used tree space prototypes to sparsen a tree \citep{TanHookerEtAl2016} or rectified decision trees that use hard and soft labels \citep{BaiLiEtAl2019}. \citet{FriedmanPopescu2008} followed a different concept and proposed a linear model to extract rules from a tree ensemble, which can then be used to rebuilt a single tree. Similarly, \citet{SirikulviriyaSinthupinyo2011} focused on deducing rules from a random forest, while \citet{HaraHayashi2016} focused on rule extraction from tree ensembles via bayesian model selection, and \citet{MollasTsoumakasEtAl2019} used a local-based, path-oriented similarity metric to select rules from a tree ensemble. Recently, some works focused on directly extracting a single tree from a tree ensemble based on stabilized but yet heuristic splitting criteria \citep{ZhouHooker2016}, genetic algorithms \citep{VandewieleLannoyeEtAl2017}, or by actively sampling training points \citep{BastaniKimEtAl2017,BastaniKimEtAl2017b}. All of these works focus on the creation of sparse decision trees that remain interpretable but can be used to replace a tree ensemble while securing a similar prediction performance. \myblue{However, these approaches do not guarantee faithfulness, such that the new classifier is not guaranteed to retain the same decision function and prediction performance}.

In the field of neural networks, \myblue{related studies were done on} \textit{model compression} \citep{BuciluaCaruanaEtAl2006}. The proposed approaches often \myblue{use} knowledge distillation, i.e., using a high-capacity teacher to train a compact student with similar knowledge \citep[see, e.g.,][]{HintonVinyalsEtAl2015}. Recent works focused on creating soft decision trees from a neural network \citep{FrosstHinton2017}, decomposing the gradient in knowledge distillation \citep{FurlanelloLiptonEtAl2018}, deriving a class of models for self-explanatory neural networks \citep{MelisJaakkola2018}, or specified knowledge representations in high conv-layers for interpretable convolutional neural networks \citep{ZhangNianWuEtAl2018}. Focusing on feed-forward neural networks, \citet{FrankleCarbin2018} \myblue{proposed} pruning techniques that identify subnetworks which perform close to the original network. \citet{ClarkLuongEtAl2019} studied born-again multi task networks for natural language processing, while \citet{KisamoriYamazaki2019} focused on synthesizing an interpretable simulation model from a neural network. As neural networks are highly non-linear and even less transparent \myblue{than} tree ensembles, all of these approaches remain predominantly heuristic \myblue{and faithfulness is typically not achievable}.

\paragraph{\myblue{Optimal decision} trees.}
Since the 1990's, some works focused on constructing decision trees based on mathematical programming techniques. \citet{Bennett1992} used linear programming to construct trees with linear combination splits and showed that this technique performs better than conventional univariate split algorithms. \citet{BennettBlue1996} focused on building global optimal decision trees to avoid overfitting, while \citet{NijssenFromont2007} presented an exact algorithm to build a decision tree for specific depth, accuracy, and leaf requirements. Recently, \citet{BertsimasDunn2017} presented a mixed integer programming formulation to construct optimal classification trees. On a similar note, \citet{GuenluekKalagnanamEtAl2018} presented an integer programming approach for optimal decision trees with categorical data, and \citet{VerwerZhang2019} presented a binary linear program for optimal decision trees. \citet{HuRudinEtAl2019} presented a scalable algorithm for optimal sparse binary decision trees. While all these works show that decision trees are in general amenable to be built with optimization techniques, none of these works focused on constructing born-again trees that match the accuracy of a given tree ensemble. 

\paragraph{Summary.} Thinning problems have been studied for both tree ensembles and neural networks in order to derive interpretable classifiers that show a similar performance than the aforementioned algorithms. However, all of these works embed heuristic construction techniques or an approximative objective, such that the resulting classifiers do not guarantee a behavior and prediction performance equal to the original tree ensemble or neural network. These approaches appear to be plausible for born-again neural networks, as neural networks have highly non-linear structures that cannot be easily captured in an optimization approach. \myblue{In contrast}, work in the field of building optimal decision trees showed that the construction of decision trees is generally amenable for optimization based approaches. Nevertheless, these works focused so far on constructing sparse or optimal trees that outperform heuristically created trees, such that the question whether one could construct an optimal decision tree that serves as a born-again tree ensemble remains open. Answering this question and discussing some of its implications is the focus of our study.

\subsection{Contributions}
\label{subsec:contribution}

With this work, we revive the concept of born-again tree ensembles \myblue{and aim to construct a single ---minimum-size--- tree that faithfully reproduces the decision function of the original tree ensemble. More} specifically, our contribution is fourfold. First, we formally define the problem of constructing optimal born-again tree ensembles and prove that this problem is \myblue{NP-hard}. Second, we \myblue{highlight several properties} of this problem and \myblue{of} the resulting born-again tree. These findings allow us to develop a dynamic-programing based algorithm that solves this problem efficiently and constructs an \myblue{optimal} born-again tree out of a tree ensemble. Third, we discuss specific pruning strategies for the born-again tree that allow to reduce redundancies that cannot be identified in the original tree ensemble. Fourth, besides providing theoretical guarantees, we present numerical studies which allow to analyze the characteristics of the born-again trees in terms of interpretability and accuracy. Further, these studies show that our algorithm is amenable to a wide range of real-world data sets.

We believe that our results and the developed algorithms open a new perspective in the field of interpretable machine learning. With this approach, one can construct simple classifiers that bear all characteristics of a tree ensemble. \myblue{Besides interpretability gains, this approach casts a new light on tree ensembles and highlights new structural properties.}

\section{Fundamentals}
\label{sec:fundamentals}

In this section, we introduce some fundamental definitions. Afterwards, we discuss a worst-case bound \myblue{on} the depth of an optimal born-again tree.

\paragraph{Tree ensemble.} We define a tree ensemble $\mathcal{T}$ as a set of trees $t \in \mathcal{T}$ with weights $w_t$.
For any sample~$\mathbf{x}$, the tree ensemble returns the majority vote of its trees: $F_\mathcal{T}(\mathbf{x}) = \myblue{\textsc{Weighted-Majority}}\{(F_t(\mathbf{x}),w_t)\}_{t\in\mathcal{T}}$ (ties are broken in favor of the  smaller index).

\paragraph{Cells.} Let $H_j$ be the set of all split levels (i.e., hyperplanes) extracted from the trees for each feature~$j$.
We can \myblue{partition} the feature space $\mathbb{R}^p$ into cells $\mathcal{S}_\textsc{elem} = \{1,\dots,|H_1|+1\}\times\dots\times \{1,\dots,|H_p|+1\}$ such that each cell $\mathbf{z} = (z_1,\dots,z_p) \in \mathcal{S}_\textsc{elem}$ represents the box contained between the $(z_j-1)^\text{th}$ and $z_j^\text{th}$ hyperplanes for each feature $j \in \{1,\dots,p\}$. Cells such that $z_j=1$ (or $z_j=|H_j|+1$) extend from $-\infty$ (or to $\infty$, respectively) along dimension~$j$. We note that \myblue{the decision function of the tree ensemble $F_\mathcal{T}(\mathbf{z})$ is} constant in the interior of each cell \myblue{$\mathbf{z}$}, \myblue{allowing} us to exclusively use the hyperplanes of $\{H_j\}_{j=1}^d$ to construct \myblue{an optimal} born-again tree.

\paragraph{Regions.} We define a \emph{region} of the feature space as a pair $(\mathbf{z}^\textsc{l},\mathbf{z}^\textsc{r}) \in \mathcal{S}_\textsc{elem}^2$ such that $\mathbf{z}^\textsc{l} \leq \mathbf{z}^\textsc{r}$. Region $(\mathbf{z}^\textsc{l},\mathbf{z}^\textsc{r})$  encloses all cells $\mathbf{z}$ such that $\mathbf{z}^\textsc{l} \leq \mathbf{z} \leq \mathbf{z}^\textsc{r}$. Let $\mathcal{S}_\textsc{regions}$ be the set of all regions.
An optimal born-again tree $T$ for a region $(\mathbf{z}^\textsc{l},\mathbf{z}^\textsc{r})$ is a tree of minimal size such that $F_T(\mathbf{x}) = F_\mathcal{T}(\mathbf{x})$ within this region.

\begin{figure}[h]
\centering
	\includegraphics[width = 0.48\textwidth]{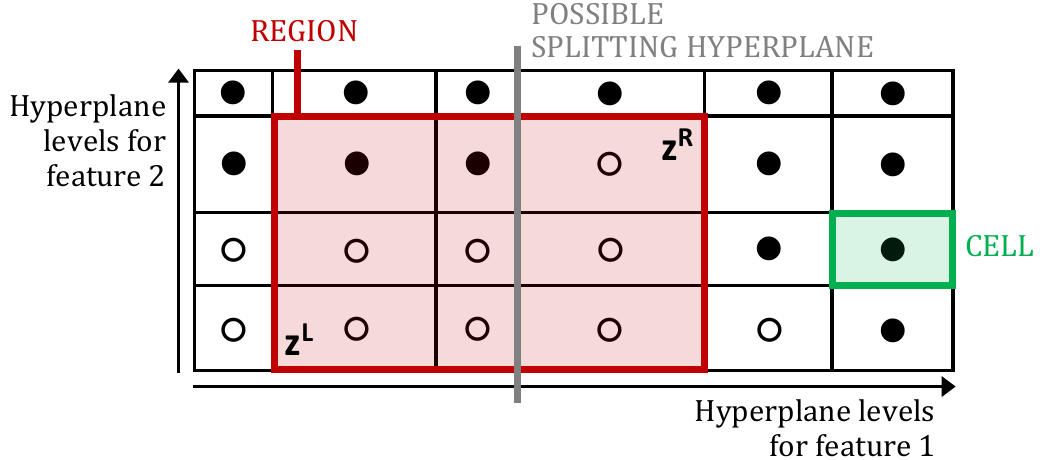}
    \vspace*{-0.45cm}
	\caption{Example of a cell, region and splitting hyperplane\label{figure-space}}
\end{figure}

Figure~\ref{figure-space} depicts a cell and a region on a two-dimensional feature space. \myblue{We also provide a more extensive example of the born-again tree generation process in the supplementary material.} 
The number of cells and regions increases rapidly with the number of hyperplanes and features, formally:
\begin{equation}
|\mathcal{S}_\textsc{elem}| = \prod_{j=1}^p (|H_j|+1)
\end{equation}
\begin{equation}
\label{count-regions}
|\mathcal{S}_\textsc{region}| = \prod_{j=1}^p \frac{(|H_j|+1)(|H_j|+2)}{2}.
\end{equation}

Moreover, Theorem~\ref{theorem:depth} gives initial bounds on the size of the born-again decision tree.

\begin{theorem}
\label{theorem:depth}
The depth of \myblue{an optimal} born-again tree~$T$ satisfies 
$\Phi(T) \leq \sum_{t \in \mathcal{T}} \Phi(t)$, where $\Phi(t)$ represents the depth of a tree $t$. This bound is tight.
\end{theorem}

This bound corresponds to a worst case behavior which is usually attained only on purposely-designed pathological cases. As highlighted in our computational experiments, the average tree depth remains generally lower than this analytical worst case. Beyond interpretability benefits, the tree depth represents the number of sequential operations (hyperplane comparisons) needed to determine the class of a given sample during the test stage. Therefore, an optimized born-again tree is not only more interpretable, but it also requires less test effort, with useful applications for classification in embarked systems, typically occurring within limited time and processing budgets.

\section{Methodology}
\label{sec:methodology}

In this section, we introduce a dynamic programming (DP) algorithm \myblue{which optimally} solves Problem~\ref{prob:bornagain} for many \myblue{data sets} of practical interest. Let $\Phi(\mathbf{z}^\textsc{l},\mathbf{z}^\textsc{r})$ be the depth of an optimal born-again decision tree for a region $(\mathbf{z}^\textsc{l},\mathbf{z}^\textsc{r}) \in \mathcal{S}_\textsc{region}$. We can then limit the search to optimal born-again trees whose left and right sub-trees represent optimal born-again trees for the respective sub-regions. Hence, we can recursively decompose a larger problem into subproblems using
\begin{align}
&\Phi(\mathbf{z}^\textsc{l},\mathbf{z}^\textsc{r}) = \label{eq-DP0} \\
&\begin{cases} 0 \ \text{ if } \ \textsc{id}(\mathbf{z}^\textsc{l},\mathbf{z}^\textsc{r}) \\
\min\limits_{1 \leq j \leq p} \left\{ \min\limits_{z^\textsc{l}_j \leq l < z^\textsc{r}_j} \left\{ 1+ \max \{ 
\Phi(\mathbf{z}^\textsc{l},\mathbf{z}^\textsc{r}_{jl}),
\Phi(\mathbf{z}^\textsc{l}_{jl},\mathbf{z}^\textsc{r})
\} \right\} \right\}, \nonumber
\end{cases}
\end{align}
in which $\textsc{id}(\mathbf{z}^\textsc{l},\mathbf{z}^\textsc{r})$ takes value \textsc{True} if and only if all cells $\mathbf{z}$ such that $\mathbf{z}^\textsc{l} \leq \mathbf{z} \leq \mathbf{z}^\textsc{r}$ admit the same weighted majority class. In this equation, $\mathbf{z}^\textsc{r}_{jl}=\mathbf{z}^\textsc{r}+\mathbf{e}_j(l - z^\textsc{r}_j)$ represents the ``top right'' corner of the left region obtained in the subdivision, and $\mathbf{z}^\textsc{l}_{jl}=\mathbf{z}^\textsc{l}+\mathbf{e}_j(l + 1 - z^\textsc{l}_j)$ is the ``bottom left'' corner of the right region obtained in the subdivision.

While Equation~(\ref{eq-DP0}) bears the main rationale of our algorithm, it suffers in its basic state from two main weaknesses that prevent its translation into an efficient algorithm: firstly, each verification of the first condition (i.e., the base case) requires evaluating whether $\textsc{id}(\mathbf{z}^\textsc{l},\mathbf{z}^\textsc{r})$ is true and possibly requires the evaluation of the majority class on an exponential number of cells \emph{if done brute force}. Secondly, the recursive call considers all possible hyperplanes within the region to find the minimum over $j \in \{1,\dots,p\}$ and $z^\textsc{l}_j \leq l < z^\textsc{r}_j$. In the following, we propose strategies to mitigate both drawbacks.

To avoid the evaluation of $\textsc{id}(\mathbf{z}^\textsc{l},\mathbf{z}^\textsc{r})$ by inspection, we integrate this evaluation within the recursion to profit from the memory structures of the DP algorithm.
With these changes, the recursion becomes:
\begin{align}
&\Phi(\mathbf{z}^\textsc{l},\mathbf{z}^\textsc{r}) = \label{eq-DP1} \\
&
\hspace*{-0.05cm} \min\limits_{j} \hspace*{-0.05cm} \left\{ \hspace*{-0.05cm} \min\limits_{z^\textsc{l}_j \leq l < z^\textsc{r}_j} \hspace*{-0.1cm} \left\{ \mathds{1}_{jl}(\mathbf{z}^\textsc{l},\mathbf{z}^\textsc{r})+ \max \{ 
\Phi(\mathbf{z}^\textsc{l},\mathbf{z}^\textsc{r}_{jl}),
\Phi(\mathbf{z}^\textsc{l}_{jl},\mathbf{z}^\textsc{r})
\} \right\} \hspace*{-0.08cm} \right\} \nonumber \\
&\text{where} \ \mathds{1}_{jl}(\mathbf{z}^\textsc{l},\mathbf{z}^\textsc{r}) =
\begin{cases}
0\hspace*{-0.08cm}&
\begin{aligned}
&\text{if\hspace*{-0.2cm}}&&\Phi(\mathbf{z}^\textsc{l},\mathbf{z}^\textsc{r}_{jl}) =
\Phi(\mathbf{z}^\textsc{l}_{jl},\mathbf{z}^\textsc{r}) = 0 \\
&\text{and\hspace*{-0.2cm}}&&F_\mathcal{T}(\mathbf{z}^\textsc{l}) = F_\mathcal{T}(\mathbf{z}^\textsc{r});
\end{aligned}
 \\
1\hspace*{-0.08cm}& \text{otherwise.} 
\end{cases}
  \nonumber 
\end{align}

To limit the number of recursive calls, \myblue{we can filter out for each dimension $j$ any hyperplane $l \in \{1,\dots,|H_j|\}$ such that $F_\mathcal{T}(\mathbf{z}) = F_\mathcal{T}(\mathbf{z}+\mathbf{e}_j)$ for all $\mathbf{z}$ such that \mbox{$\mathbf{z}_j=l$}, and exploit} two additional properties of the problem.

\begin{theorem}
\label{theorem:monotonicity}
Let $(\mathbf{z}^\textsc{l},\mathbf{z}^\textsc{r})$ and $(\mathbf{\bar{z}}^\textsc{l},\mathbf{\bar{z}}^\textsc{r})$ be two regions such that $\mathbf{z}^\textsc{l} \leq \mathbf{\bar{z}}^\textsc{l} \leq \mathbf{\bar{z}}^\textsc{r} \leq \mathbf{z}^\textsc{r}$, then $\Phi(\mathbf{\bar{z}}^\textsc{l},\mathbf{\bar{z}}^\textsc{r}) \leq \Phi(\mathbf{z}^\textsc{l},\mathbf{z}^\textsc{r}).$ 
\end{theorem}

Theorem~\ref{theorem:monotonicity} follows from the fact that any feasible tree satisfying $F_T(\mathbf{x}) = F_\mathcal{T}(\mathbf{x})$ on a region $(\mathbf{z}^\textsc{l},\mathbf{z}^\textsc{r})$ also satisfies this condition for any subregion $(\mathbf{\bar{z}}^\textsc{l},\mathbf{\bar{z}}^\textsc{r})$.
Therefore, $\Phi(\mathbf{\bar{z}}^\textsc{l},\mathbf{\bar{z}}^\textsc{r})$ constitutes a lower bound of $\Phi(\mathbf{z}^\textsc{l},\mathbf{z}^\textsc{r})$. Combining this bound with Equation~(\ref{eq-DP1}), we get
\begin{equation}
\begin{aligned}
\max &\{\Phi(\mathbf{z}^\textsc{l},\mathbf{z}^\textsc{r}_{jl}),\Phi(\mathbf{z}^\textsc{l}_{jl},\mathbf{z}^\textsc{r}) \} \\ 
&\leq \Phi(\mathbf{z}^\textsc{l},\mathbf{z}^\textsc{r}) \nonumber \\
&\leq \mathds{1}_{jl}(\mathbf{z}^\textsc{l},\mathbf{z}^\textsc{r}) +\max \{
\Phi(\mathbf{z}^\textsc{l},\mathbf{z}^\textsc{r}_{jl}),
\Phi(\mathbf{z}^\textsc{l}_{jl},\mathbf{z}^\textsc{r})\}
\end{aligned}\nonumber
\end{equation}
for each $j \in \{1,\dots,p\}$ and $z^\textsc{l}_j \leq l < z^\textsc{r}_j$.

This result will be fundamental to use bounding techniques and therefore save numerous recursions during the DP algorithm. 
With Theorem~\ref{theorem:binarysearch}, we can further reduce the number of candidates in each recursion.

\begin{theorem}
\label{theorem:binarysearch}
Let $j \in \{1,\dots,p\}$ and $l \in \{z^\textsc{l}_j,\dots,z^\textsc{r}_j-1\}$.
\begin{itemize}[nosep,leftmargin=*]
\item If $\Phi(\mathbf{z}^\textsc{l},\mathbf{z}^\textsc{r}_{jl}) \geq \Phi(\mathbf{z}^\textsc{l}_{jl},\mathbf{z}^\textsc{r})$ then $\forall l'>l$
\begin{align}
&\mathds{1}_{jl}(\mathbf{z}^\textsc{l},\mathbf{z}^\textsc{r})+ \max \{ 
\Phi(\mathbf{z}^\textsc{l},\mathbf{z}^\textsc{r}_{jl}),
\Phi(\mathbf{z}^\textsc{l}_{jl},\mathbf{z}^\textsc{r})
\} \nonumber \\
\leq \ &\mathds{1}_{jl'}(\mathbf{z}^\textsc{l},\mathbf{z}^\textsc{r})+ \max \{ 
\Phi(\mathbf{z}^\textsc{l},\mathbf{z}^\textsc{r}_{jl'}),
\Phi(\mathbf{z}^\textsc{l}_{jl'},\mathbf{z}^\textsc{r})
\} \nonumber
\end{align}
\item If $\Phi(\mathbf{z}^\textsc{l},\mathbf{z}^\textsc{r}_{jl}) \leq \Phi(\mathbf{z}^\textsc{l}_{jl},\mathbf{z}^\textsc{r})$ then $\forall l'<l$
\begin{align}
&\mathds{1}_{jl}(\mathbf{z}^\textsc{l},\mathbf{z}^\textsc{r})+ \max \{ 
\Phi(\mathbf{z}^\textsc{l},\mathbf{z}^\textsc{r}_{jl}),
\Phi(\mathbf{z}^\textsc{l}_{jl},\mathbf{z}^\textsc{r})
\} \nonumber \\
\leq \ &\mathds{1}_{jl'}(\mathbf{z}^\textsc{l},\mathbf{z}^\textsc{r})+ \max \{ 
\Phi(\mathbf{z}^\textsc{l},\mathbf{z}^\textsc{r}_{jl'}),
\Phi(\mathbf{z}^\textsc{l}_{jl'},\mathbf{z}^\textsc{r})
\}. \nonumber
\end{align}
\end{itemize}
\end{theorem}

Based on Theorem~\ref{theorem:binarysearch}, we can discard all hyperplane levels $l' > l$ in Equation~(\ref{eq-DP1}) if $\Phi(\mathbf{z}^\textsc{l},\mathbf{z}^\textsc{r}_{jl}) \geq \Phi(\mathbf{z}^\textsc{l}_{jl},\mathbf{z}^\textsc{r})$. The same argument holds when $\Phi(\mathbf{z}^\textsc{l},\mathbf{z}^\textsc{r}_{jl}) \leq \Phi(\mathbf{z}^\textsc{l}_{jl},\mathbf{z}^\textsc{r})$ with $l'<l$. We note that the two cases of Theorem~\ref{theorem:binarysearch} are not mutually exclusive. No other recursive call is needed for the considered feature when an equality occurs. Otherwise, at least one case holds, allowing us to search the range $l \in \{z^\textsc{l}_j,\dots,z^\textsc{r}_j-1\}$ in Equation~(\ref{eq-DP1}) by binary search with only $O(\log(z^\textsc{r}_j-z^\textsc{l}_j))$ subproblem calls.

\paragraph{General algorithm structure.}
The DP algorithm presented in Algorithm~\ref{algo-DP} capitalizes upon all the aforementioned properties. It is initially launched on the region representing the complete feature space, by calling  \textsc{Born-Again}($\mathbf{z}^\textsc{l},\mathbf{z}^\textsc{r}$) with $\mathbf{z}^\textsc{l} = (1,\dots,1)^\intercal$ and $\mathbf{z}^\textsc{r} = (|H_1|+1,\dots,|H_p|+1)^\intercal$. 

Firstly, the algorithm checks whether it attained a base case in which the region ($\mathbf{z}^\textsc{l},\mathbf{z}^\textsc{r}$) is restricted to a single cell (Line~1). If this is the case, it returns an optimal depth of zero corresponding to a single leaf, otherwise it tests whether the result of the current subproblem defined by region $(\mathbf{z}^\textsc{l},\mathbf{z}^\textsc{r})$ is not yet in the DP memory (Line~2). If this is the case, it directly returns the known result.

Past these conditions, the algorithm starts enumerating possible splits and opening recursions to find the minimum of Equation~(\ref{eq-DP1}). By Theorem~\ref{theorem:binarysearch} and the related discussions, it can use a binary search for each feature to save many possible evaluations (Lines 9 and 10). By Theorem~\ref{theorem:monotonicity}, the exploitation of lower and upper bounds on the optimal solution value (Lines 7, 9, 20, and 21) allows to stop the iterative search whenever no improving solution can exist. Finally, the special case of Lines 13 and 14 covers the case in which $\Phi(\mathbf{z}^\textsc{l},\mathbf{z}^\textsc{r}_{jl}) = \Phi(\mathbf{z}^\textsc{l}_{jl},\mathbf{z}^\textsc{r}) = 0$ and $F_\mathcal{T}(\mathbf{z}^\textsc{l}) = F_\mathcal{T}(\mathbf{z}^\textsc{r})$, corresponding to a homogeneous region in which all cells have the same majority class. As usual in DP approaches, our algorithm memorizes the solutions of sub-problems and reuses them in future calls (Lines 15, 17, and~26).

We observe that this algorithm maintains the optimal solution of each subproblem in memory, but not the solution itself in order to reduce memory consumption. Retrieving the solution after completing the DP can be done with a simple inspection of the final states and solutions, \myblue{as detailed in the supplementary material}. 

The maximum number of possible regions is $|\mathcal{S}_\textsc{region}| = \prod_j \frac{(|H_j|+1)(|H_j|+2)}{2}$ (Equation~\ref{count-regions}) and each call to \textsc{Born-Again} takes up to $O(\sum_j \log |H_j|)$ elementary operations due to Theorem~\ref{theorem:binarysearch}, leading to a worst-case complexity of $O( |\mathcal{S}_\textsc{region}| \sum_j \log |H_j|)$ time for the overall recursive algorithm. Such an exponential complexity is expectable for an NP-hard problem. Still, as observed in our experiments, the number of regions explored with the bounding strategies is much smaller in practice than the theoretical worst case.

\begin{algorithm}[htbp]
\caption{\textsc{Born-Again}($\mathbf{z}^\textsc{l},\mathbf{z}^\textsc{r})$}
\label{algo-DP}
\begin{algorithmic}[1]
\STATE \textbf{if} $(\mathbf{z}^\textsc{l} = \mathbf{z}^\textsc{r})$ \textbf{return} 0
\IF{$(\mathbf{z}^\textsc{l},\mathbf{z}^\textsc{r})$ exists in memory}
\STATE \textbf{return} $\textsc{Memory}(\mathbf{z}^\textsc{l},\mathbf{z}^\textsc{r})$
\ENDIF
\STATE $\textsc{UB} \gets \infty$
\STATE $\textsc{LB} \gets 0$
\FOR{$j=1$ {\bfseries to} $p$ \text{ and } $\textsc{LB} < \textsc{UB}$}
	\STATE $(\textsc{Low},\textsc{Up}) \gets (z_j^\textsc{l},\myblue{z}_j^\textsc{r})$
	\WHILE{$\textsc{Low} < \textsc{Up}$ \text{ and } $\textsc{LB} < \textsc{UB}$}
		\STATE $l \gets \lfloor (\textsc{Low}+\textsc{Up})/2 \rfloor$
		\STATE $\Phi_1 \gets \textsc{Born-Again}(\mathbf{z}^\textsc{l},\mathbf{z}^\textsc{r}+\mathbf{e}_j(l - z^\textsc{r}_j))$
		\STATE $\Phi_2 \gets \textsc{Born-Again}(\mathbf{z}^\textsc{l}+\mathbf{e}_j(l + 1 - z^\textsc{l}_j),\mathbf{z}^\textsc{r})$
		\IF {$(\Phi_1 = 0)$ \text{and} $(\Phi_2 = 0)$}
		\STATE \textbf{if} $f(\mathbf{z}^\textsc{l},\mathcal{T}) = f(\mathbf{z}^\textsc{r},\mathcal{T})$ \textbf{then} 
		\STATE \hspace*{0.5cm} \textsc{Memorize}(($\mathbf{z}^\textsc{l},\mathbf{z}^\textsc{r}),0$) \textbf{and return} 0
        \STATE \myblue{\textbf{else}}
		\STATE \hspace*{0.5cm} \textsc{Memorize}(($\mathbf{z}^\textsc{l},\mathbf{z}^\textsc{r}),1$) \textbf{and return} 1
		\STATE \textbf{end if}
		\ENDIF
	\STATE $\textsc{UB} \gets \min \{\textsc{UB},1+\max\{\Phi_1,\Phi_2\}\}$ 
	\STATE $\textsc{LB} \gets \max \{\textsc{LB},\max\{\Phi_1,\Phi_2\}\}$ 
	\STATE \textbf{if} $(\Phi_1 \geq \Phi_2)$ \textbf{then} $\textsc{Up} \gets l$
	\STATE \textbf{if} $(\Phi_1 \leq \Phi_2)$ \textbf{then} $\textsc{Low} \gets l+1$
	\ENDWHILE
\ENDFOR
	\STATE \textsc{Memorize}(($\mathbf{z}^\textsc{l},\mathbf{z}^\textsc{r}),\textsc{UB}$) \textbf{and return} UB
\end{algorithmic}
\end{algorithm}

\section{Computational Experiments}
\label{sec:experiments}

The goal of our computational experiments is \myblue{fourfold}:
\begin{enumerate}[nosep]
\item Evaluating the computational performance of the proposed DP algorithm as a function of the data set characteristics, e.g., the size metric in use, the number of trees in the original ensemble, and the number of samples and features in the datasets. 
\item \myblue{Studying} the structure \myblue{and complexity} of the born-again trees for different size metrics.
\item Measuring the impact of a simple pruning strategy applied on the resulting born-again trees.
\item \myblue{Proposing and evaluating a fast heuristic algorithm to find faithful born-again trees.}
\end{enumerate}

The DP algorithm was implemented in \textsc{C++} and compiled with \textsc{gcc} 9.2.0 using flag \textsc{-O3}, whereas the original random forests were generated in Python (using scikit-learn v0.22.1). All our experiments were run on a single thread of an Intel(R) Xeon(R) CPU E5-2620v4 2.10GHz, with 128\myblue{GB} of available RAM, running CentOS v7.7. In the remainder of this section, we discuss the preparation of the data and then describe each experiment. \myblue{Detailed computational results, data, and source codes are available in the supplementary material and at the following address: \url{https://github.com/vidalt/BA-Trees}.}

\subsection{Data Preparation}
\label{exp:data}

We focus on a set of six datasets from the UCI machine learning repository [UCI] and from previous work by \myblue{\citet{Smith1988} [SmithEtAl] and} \citet{HuRudinEtAl2019} [HuEtAl] for which using random forests (with ten trees) showed a significant improvement upon stand-alone CART. The characteristics of these datasets are summarized in Table~\ref{tab:datasets}: number of samples $n$, number of features $p$, number of classes $K$ and class distribution CD. To obtain discrete numerical features, we used one-hot encoding on categorical data and binned continuous features into ten ordinal scales. Then, we generated training and test samples for all data sets using a ten-fold cross validation. Finally, for each fold and each dataset, we generated a random forest composed of ten trees with a maximum depth of three (i.e., eight leaves at most), considering $p/2$ random candidate features at each split. This random forest constitutes the input to our DP algorithm.

\begin{table}[htbp]
\caption{Characteristics of the data sets}
\label{tab:datasets}
\setlength{\tabcolsep}{0.25cm}
\scalebox{0.85}
{
	\begin{tabular}{lrrrrr}
		\toprule
		\multicolumn{1}{l}{Data set} & \multicolumn{1}{r}{$n$} & \multicolumn{1}{r}{$p$} & \multicolumn{1}{r}{$K$} & \multicolumn{1}{r}{CD} & \multicolumn{1}{r}{Src.} \\
		\midrule
BC: Breast-Cancer & 683    & 9      & 2      & 65-35      & UCI \\
CP: COMPAS & 6907    & 12      & 2      & 54-46      & HuEtAl \\
FI: FICO  &10459     & 17      & 2      & 52-48      & HuEtAl \\
HT: HTRU2 & 17898      & 8      & 2      & 91-9      & UCI \\
PD: Pima-Diabetes &  768     & 8      & 2      & 65-35      & SmithEtAl \\
SE: Seeds & 210      & 7      & 3      & 33-33-33      & UCI \\
		\bottomrule
	\end{tabular}%
}
\end{table}%

\subsection{Computational Effort}
\label{exp:effort}

\begin{figure}
\centering
	\includegraphics[width=0.4\textwidth]{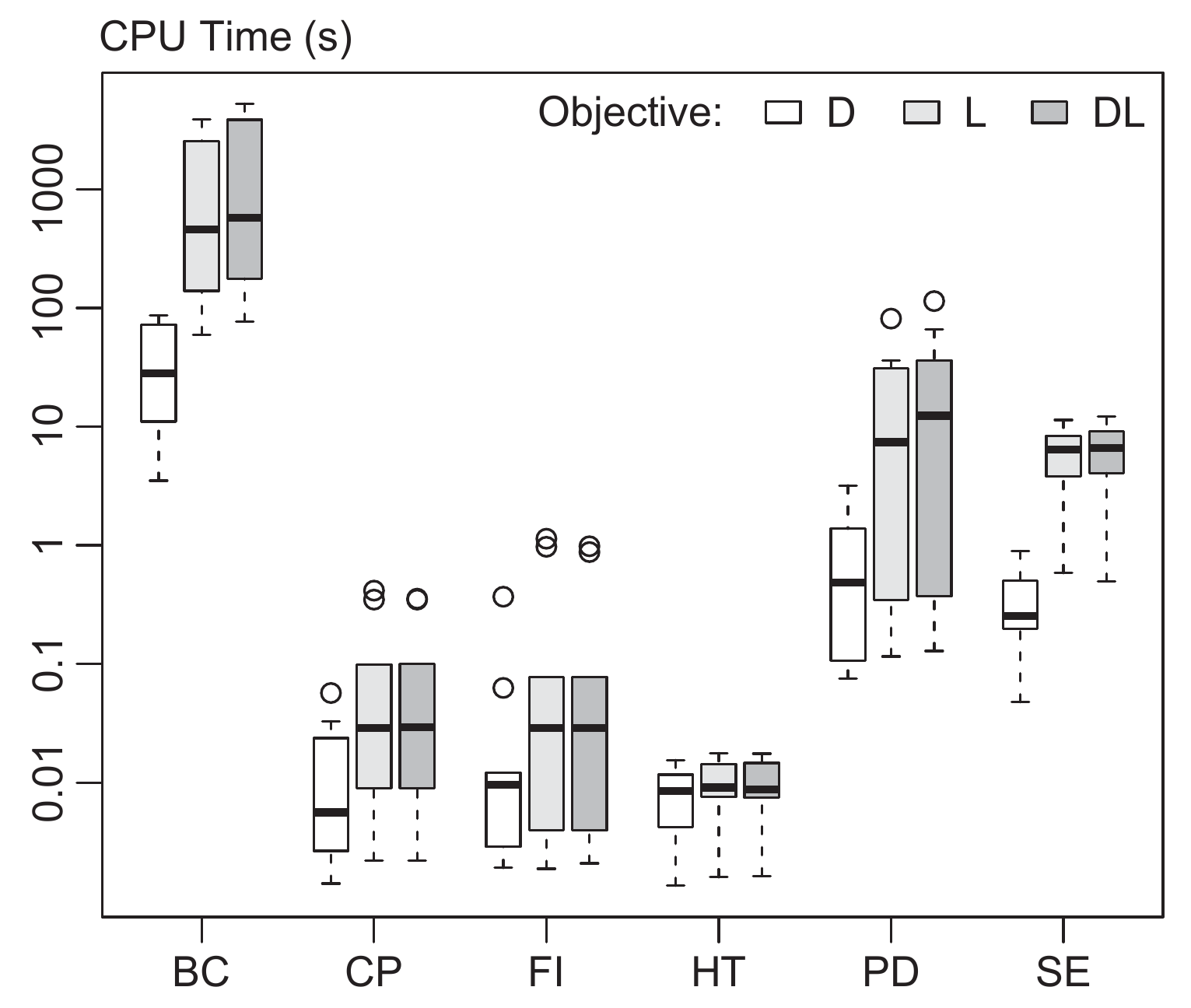}
\caption{Computational times to construct a born-again tree from a random forest with 10 trees and depth 3, for each objective (D/L/DL) and data set}
\label{fig:comptimes}
\end{figure}

In a first analysis, we evaluate the computational time of Algorithm~\ref{algo-DP} for different data sets and size metrics. Figure~\ref{fig:comptimes} reports the results of this experiment as a box-whisker plot, in which each box corresponds to ten runs (one for each fold) and the whiskers extend to $1.5$ times the interquartile range. Any sample beyond this limit is reported as outlier and noted with a ``$\circ$''. D denotes a depth-minimization objective, whereas L refers to the minimization of the number of \myblue{leaves}, and DL refers to the hierarchical objective which prioritizes the smallest depth, and then the smallest number of leaves. As can be seen, constructing a born-again tree with objective D yields significantly lower computational times compared to using objectives L and DL. Indeed, the binary search technique resulting from Theorem~\ref{theorem:binarysearch} only applies to objective D, leading to a reduced number of recursive calls in this case compared to the other algorithms.

\begin{figure*}[htbp]
\centering
\includegraphics[width = 0.31\textwidth]{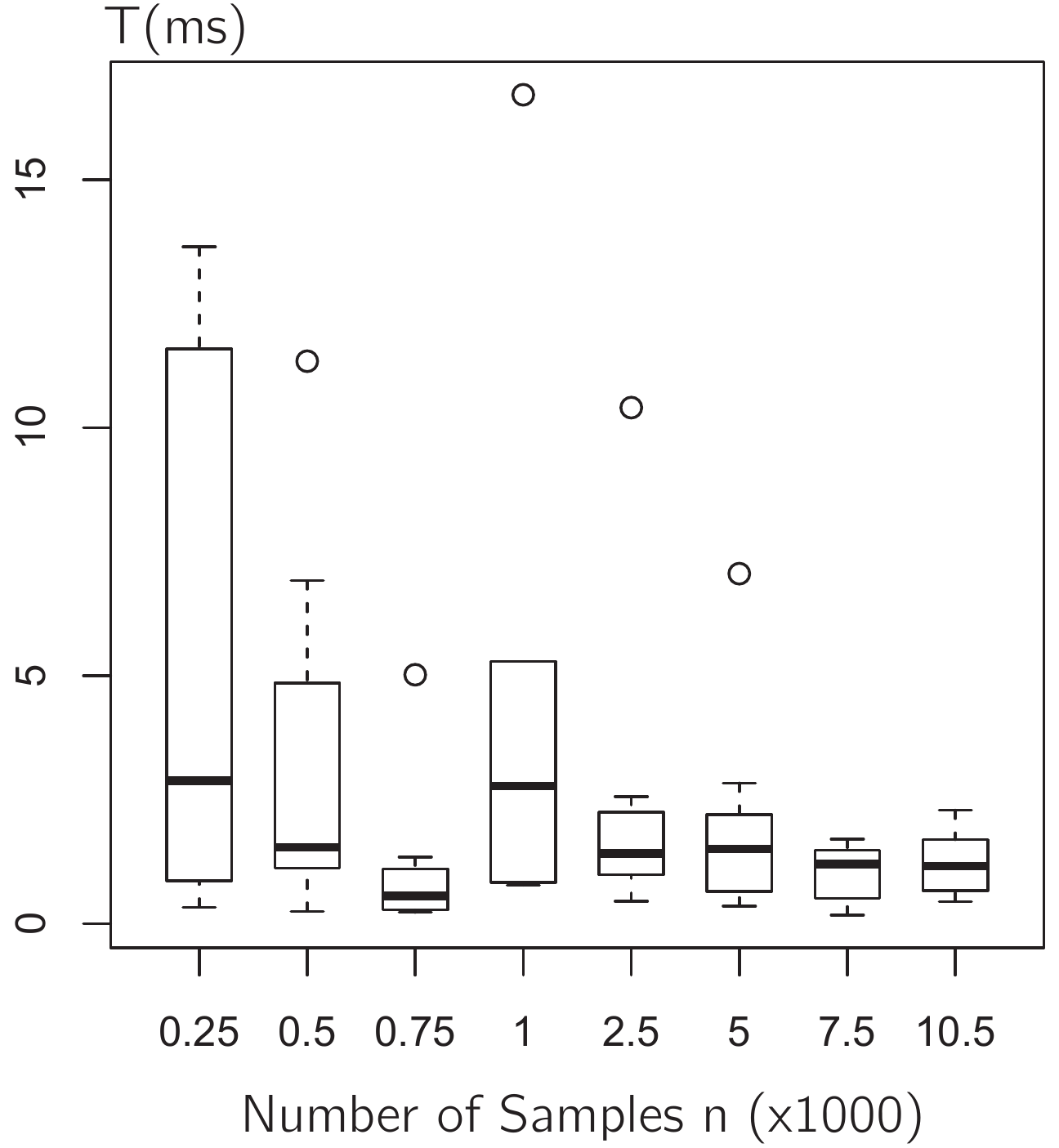}
\hspace*{0.4cm}
\includegraphics[width = 0.31\textwidth]{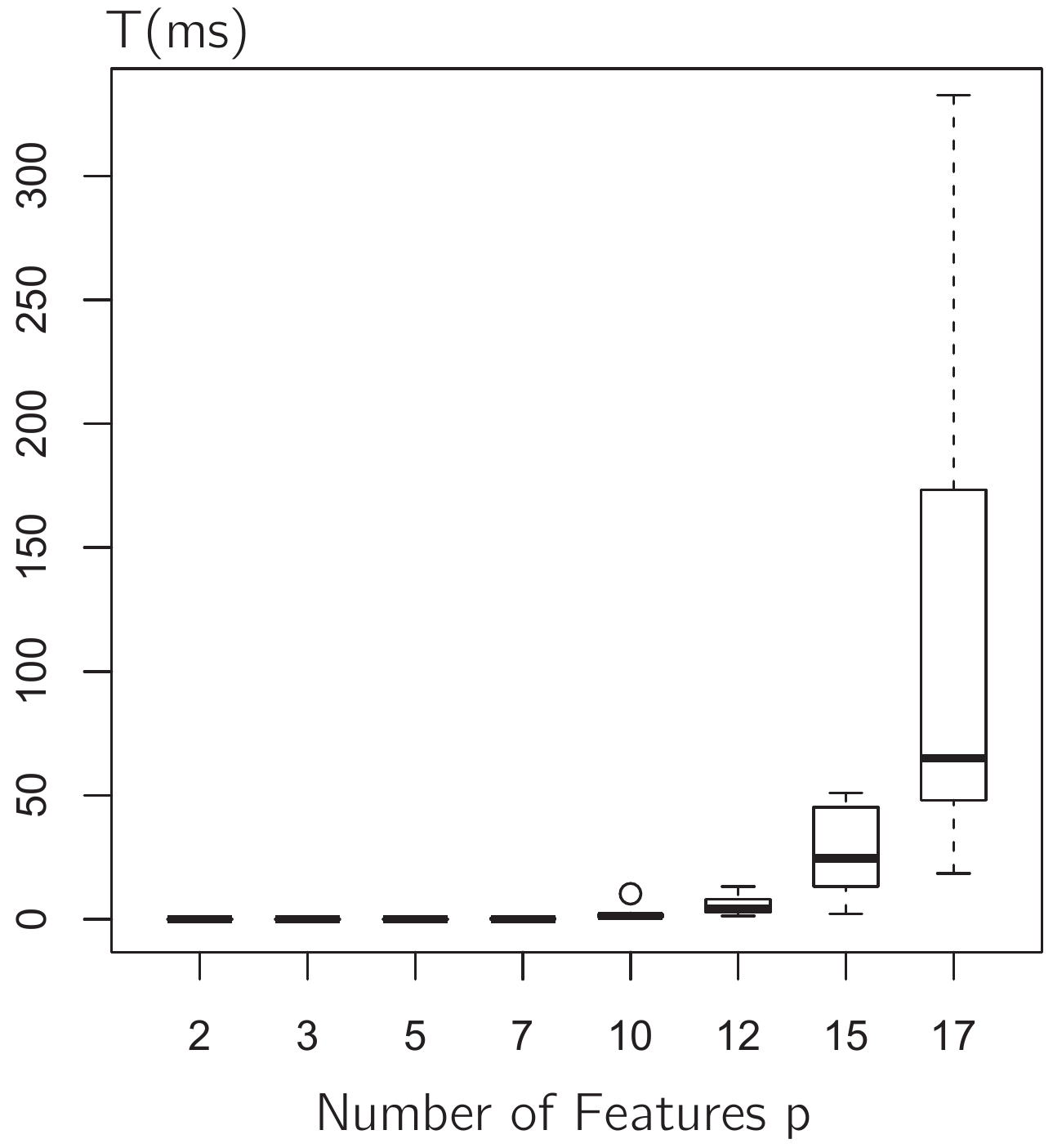}
\hspace*{0.4cm}
\includegraphics[width = 0.31\textwidth]{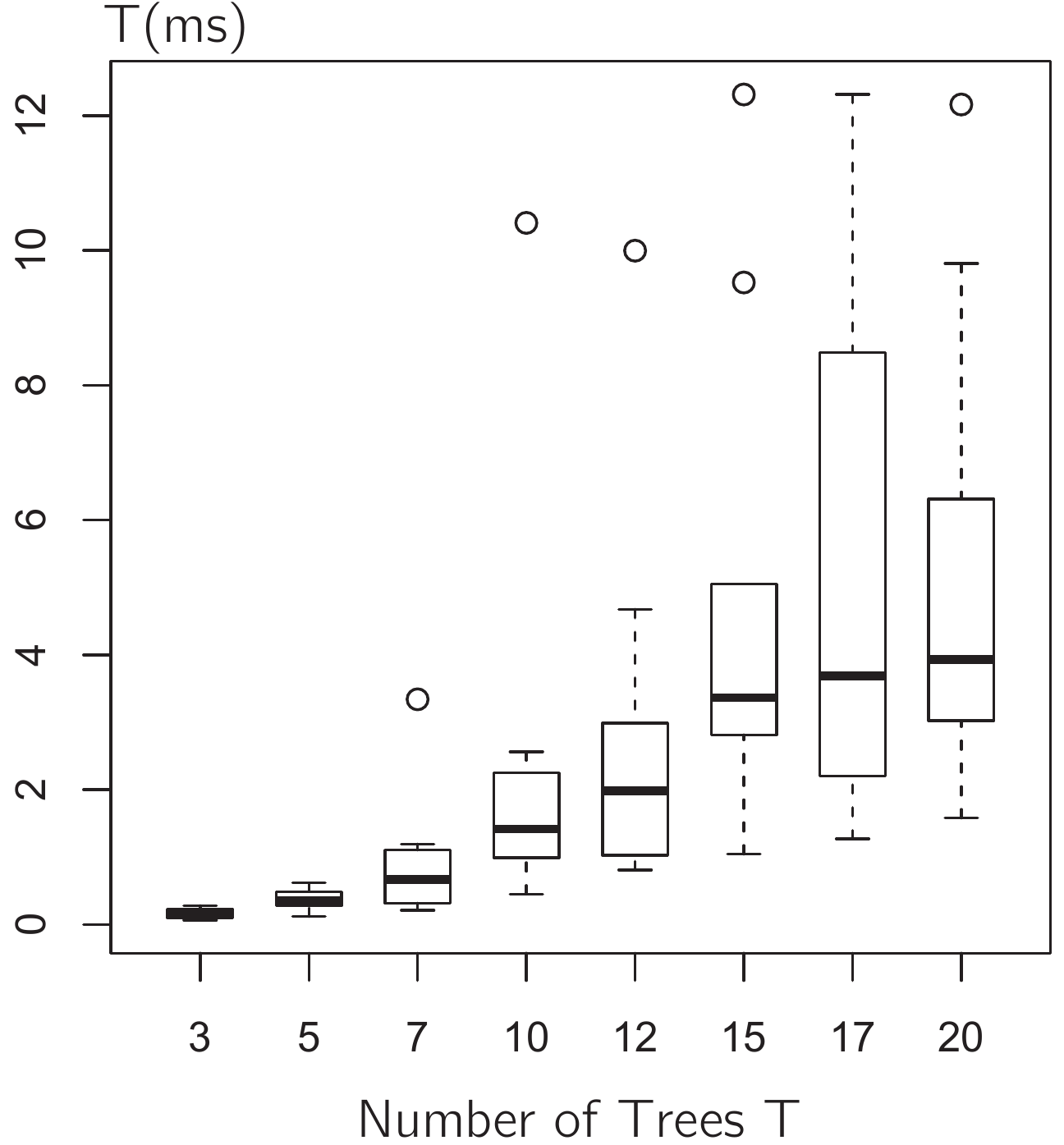}\vspace*{-0.1cm}
\caption{Growth of the computational time (in milliseconds) of Algorithm~\ref{algo-DP} as a function of the number of samples, features and trees. In each experiment, the parameters which are not under scrutiny are fixed to their baseline values of $n = 2.5 \times 10^3$, $p = 10$ and $T = 10$.}
\label{fig:scaling-subset}
\end{figure*}

In our second analysis, we focus on the FICO case and randomly extract subsets of samples and features to produce smaller data sets. We then measure the computational effort of Algorithm~\ref{algo-DP} for metric D (depth optimization) as a function of the number of features \mbox{($p \in \{2,3,5,7,10,12,15,17\}$)}, the number of samples \mbox{($n \in \{250,500,750,1000,2500,5000,7500,10459\}$)}, and the number of trees in the original random forest ($\text{T} \in \{3,5,7,10,12,15,17,20\})$. Figure~\ref{fig:scaling-subset} reports the results of this experiment. Each boxplot corresponds to ten runs, one for each fold.

We observe that the computational time of the DP algorithm is strongly driven by the number of features, with an exponential growth relative to this parameter. This result is in line with the complexity analysis of Section \ref{sec:methodology}. The number of trees influences the computational time significantly less. Surprisingly, the computational effort of the algorithm actually decreases with the number of samples. This is due to the fact that with more sample information, the decisions of the individual trees of the random forest are less varied, leading to fewer distinct \myblue{hyperplanes} and therefore \myblue{to} fewer possible states in the \myblue{DP}.

\subsection{Complexity of the Born-Again Trees}
\label{exp:structure}

We now analyze the depth and number of leaves of the born-again trees for different objective functions and datasets in Table~\ref{tab:depth-leaves-BA}.

\begin{table}[htbp]
\caption{Depth and number of leaves of the born-again trees}
\label{tab:depth-leaves-BA}
\setlength{\tabcolsep}{0.15cm}
\scalebox{0.84}
{
	\begin{tabular}{lrrrrrr}
		\toprule
& \multicolumn{2}{c}{D} & \multicolumn{2}{c}{L} & \multicolumn{2}{c}{DL} \\
\multicolumn{1}{c}{Data set} & \multicolumn{1}{c}{Depth} & \multicolumn{1}{c}{\# Leaves} & \multicolumn{1}{c}{Depth} & \multicolumn{1}{c}{\# Leaves} & \multicolumn{1}{c}{Depth} & \multicolumn{1}{c}{\# Leaves} \\
		\midrule
BC&12.5&2279.4&18.0&890.1&12.5&1042.3\\
CP&8.9&119.9&8.9&37.1&8.9&37.1\\
FI&8.6&71.3&8.6&39.2&8.6&39.2\\
HT&6.0&20.2&6.3&11.9&6.0&12.0\\
PD&9.6&460.1&15.0&169.7&9.6&206.7\\
SE&10.2&450.9&13.8&214.6&10.2&261.0\\
\midrule
Avg.&9.3&567.0&11.8&227.1&9.3&266.4\\
\bottomrule
	\end{tabular}%
}
\end{table}%

As can be seen, the different objectives can significantly influence the outcome of the algorithm. For several \myblue{data sets}, the optimal depth of the born-again tree is reached with any objective, as an indirect consequence of the minimization of the number of leaves. In other cases, however, prioritizing the minimization of the number of leaves may generate 50\% deeper trees for some data sets (e.g., PD). The hierarchical objective DL succeeds in combining the benefits of both objectives. It generates a tree with minimum depth \myblue{and} with a number of leaves which is usually close to the optimal one from objective L.

\subsection{Post-Pruned Born-Again Trees}
\label{exp:pruning}

Per definition, the born-again tree reproduces the same exact behavior as the majority class of the original ensemble classifier \emph{on all regions} of the feature space $\mathcal{X}$. Yet, some regions of $\mathcal{X}$ may not contain any training sample, either due to data scarcity or simply due to incompatible feature values (e.g., ``sex = \textsc{male}'' and ``pregnancy = \textsc{true}''). These regions may also have non-homogeneous majority classes from the tree ensemble viewpoint due to the combinations of decisions from multiple trees. The born-again tree, however, is agnostic to this situation and imitates the original classification within all the regions, leading to some splits which are mere artifacts of the ensemble's behavior but never used for classification.

\begin{figure*}[htb]
\centering
\includegraphics[scale = 0.35]{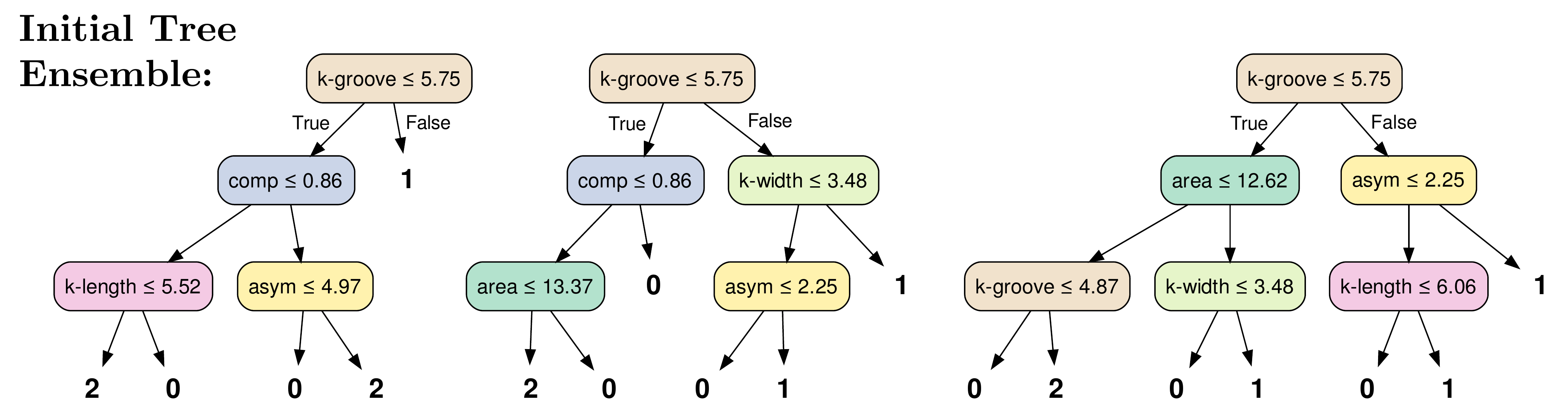}

\begin{minipage}{0.95\textwidth}
\includegraphics[scale = 0.35]{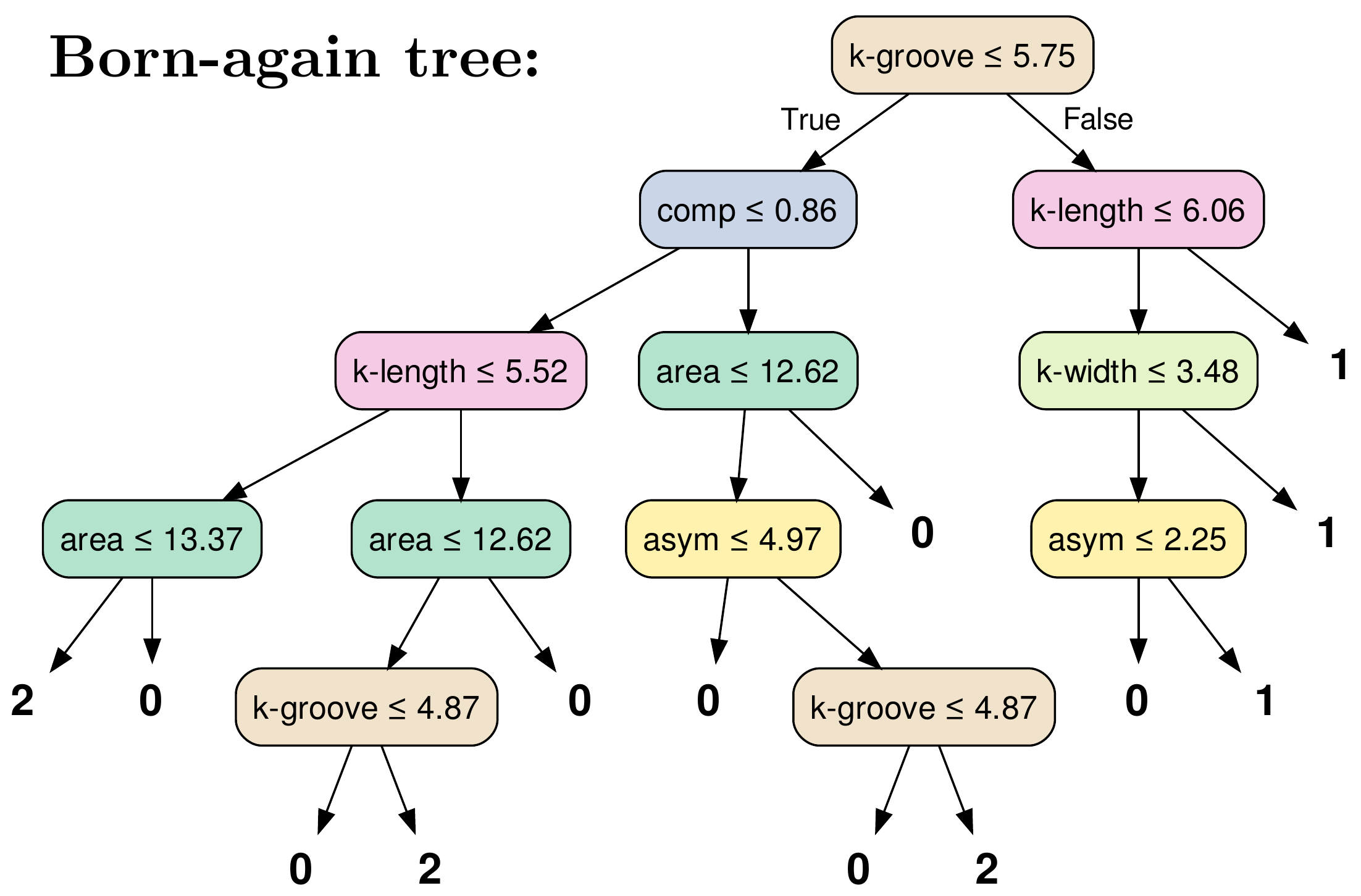}
\hfill
\includegraphics[scale = 0.45]{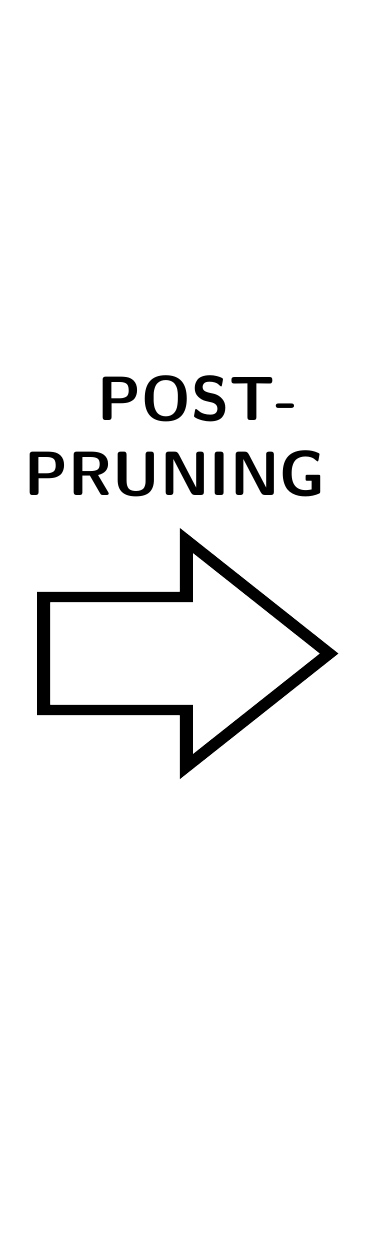}
\hfill
\includegraphics[scale = 0.35]{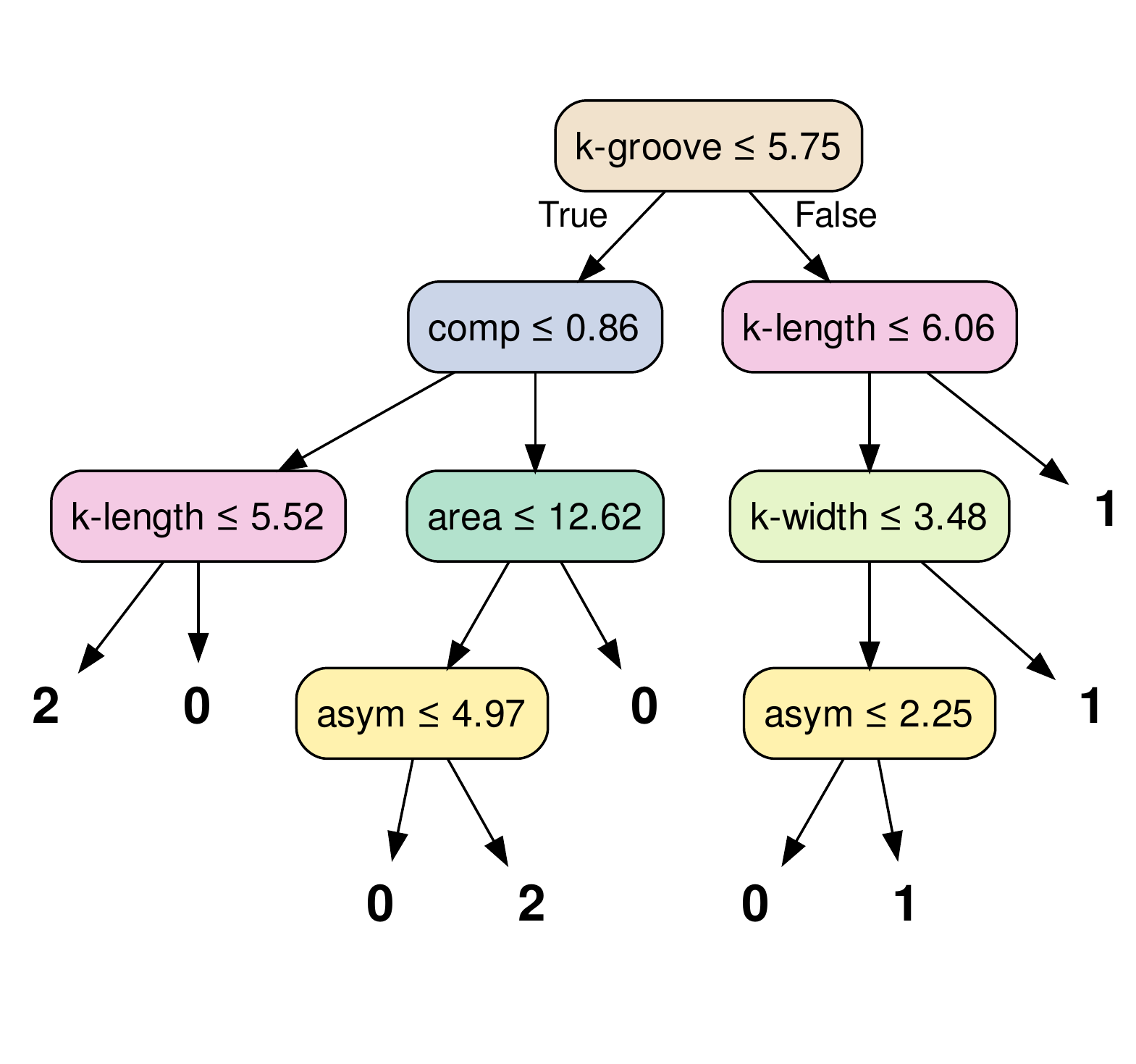}
\end{minipage}
\caption{Example of a post-pruned born-again tree on the \emph{Seeds} data set}
\label{fig:example-process}
\end{figure*}

To circumvent this issue, we suggest to apply a simple post-pruning step to eliminate inexpressive tree sub-regions. We therefore verify, from bottom to top, whether both sides of each split contain at least one training sample. Any split which does not fulfill this condition is pruned and replaced by the child node of the branch that contains samples. The complete generation process, from the original random forest to the \emph{pruned} born-again tree is illustrated in Figure~\ref{fig:example-process}. In this simple example, it is noteworthy that the born-again tree uses an optimal split at the root node which is different from all root splits in the ensemble. We also clearly observe the role of the post-pruning step, which contributes to eliminate \myblue{a significant part} of the tree.

To observe the impact of the post-pruning on a larger range of datasets, Table~\ref{tab:performance-size} reports the total number of leaves of the random forests, as well as the average depth and number of leaves of the born-again trees before and after post-pruning. As previously, the results are averaged over the ten folds. As can be seen, post-pruning significantly reduces the size of the born-again trees, leading to a final number of leaves which is, on average, smaller than the total number of leaves in the original tree ensemble. This indicates a significant gain of simplicity and interpretability.

\begin{table}[htbp]
\caption{Comparison of depth and number of leaves}
\label{tab:performance-size}
\setlength{\tabcolsep}{0.15cm}
\scalebox{0.84}
{
	\begin{tabular}{lr@{\hspace*{0.45cm}}rr@{\hspace*{0.45cm}}rr}
		\toprule
& \multicolumn{1}{c}{Random Forest} & \multicolumn{2}{c}{Born-Again} & \multicolumn{2}{c}{BA + Pruning} \\
\multicolumn{1}{c}{Data set} & \multicolumn{1}{c}{\# Leaves} & \multicolumn{1}{c}{Depth} & \multicolumn{1}{c}{\# Leaves} & \multicolumn{1}{c}{Depth} & \multicolumn{1}{c}{\# Leaves} \\
		\midrule
BC&61.1&12.5&2279.4&9.1&35.9\\
CP&46.7&8.9&119.9&7.0&31.2\\
FI&47.3&8.6&71.3&6.5&15.8\\
HT&42.6&6.0&20.2&5.1&13.2\\
PD&53.7&9.6&460.1&9.4&79.0\\
SE&55.7&10.2&450.9&7.5&21.5\\
\midrule
Avg.&51.2&9.3&567.0&7.4&32.8\\
\bottomrule
	\end{tabular}%
}
\end{table}%

However, post-pruning could cause a difference of behavior between the original tree ensemble classifier and the final pruned born-again tree. To evaluate whether this filtering had any significant impact on the classification performance of the born-again tree, we finally compare the out-of-sample accuracy (Acc.) and F1 \myblue{score} of the three classifiers in Table~\ref{tab:performance-accuracy}.

\begin{table}[htbp]
\caption{Accuracy and F1 score comparison}
\label{tab:performance-accuracy}
\setlength{\tabcolsep}{0.2cm}
\scalebox{0.84}
{
	\begin{tabular}{lrr@{\hspace*{0.65cm}}rr@{\hspace*{0.65cm}}rr}
		\toprule
& \multicolumn{2}{c}{Random Forest} & \multicolumn{2}{c}{Born-Again} & \multicolumn{2}{c}{BA + Pruning} \\
\multicolumn{1}{c}{Data set} & \multicolumn{1}{c}{Acc.} & \multicolumn{1}{c}{F1} & \multicolumn{1}{c}{Acc.} & \multicolumn{1}{c}{F1} & \multicolumn{1}{c}{Acc.} & \multicolumn{1}{c}{F1} \\
		\midrule
BC&0.953&0.949&0.953&0.949&0.946&0.941\\
CP&0.660&0.650&0.660&0.650&0.660&0.650\\
FI&0.697&0.690&0.697&0.690&0.697&0.690\\
HT&0.977&0.909&0.977&0.909&0.977&0.909\\
PD&0.746&0.692&0.746&0.692&0.750&0.700\\
SE&0.790&0.479&0.790&0.479&0.790&0.481\\
\midrule
Avg.&0.804&0.728&0.804&0.728&0.803&0.729\\
\bottomrule
	\end{tabular}
}
\end{table}

\myblue{First of all, the results of Table~\ref{tab:performance-accuracy} confirm the faithfulness of our algorithm, as they verify that the prediction quality of the random forests and the born-again tree ensembles are identical. This was expected \emph{per definition} of Problem~\ref{prob:bornagain}. Furthermore, only marginal differences were observed between the out-of-sample performance of the born-again tree with pruning and the other classifiers. For the considered datasets, pruning} contributed to eliminate inexpressive regions of the tree without much impact on classification performance.

\subsection{Heuristic Born-Again Trees}
\myblue{As Problem~\ref{prob:bornagain} is NP-hard, the computational time of our algorithm will eventually increase exponentially with the number of features (see Figure~\ref{fig:scaling-subset}). This is due to the increasing number of recursions, and to the challenge of testing homogeneity for regions without exploring all cells. Indeed, even proving that a given region is homogeneous (i.e., that it contains cells of the same class) remains NP-hard, although it is solvable in practice using integer programming techniques. Accordingly, we take a first step towards scalable heuristic algorithms in the following. We therefore explain how our born-again tree construction algorithm can be modified to preserve the faithfulness guarantee while achieving only a heuristic solution in terms of size.}


\myblue{
We made the following adaptations to Algorithm~\ref{algo-DP} to derive its heuristic counterpart. For each considered region $(\mathbf{z}^\textsc{l},\mathbf{z}^\textsc{r})$, we proceed as follows.
\begin{enumerate}
	\item Instead of evaluating all possible splits and opening recursions, we randomly select $N_c = 1000$ cells in the region and pick the splitting hyperplane that maximizes the information gain.
	\item If all these cells belong to the same class, we rely on an integer programming solver to prove whether the region is homogeneous or not. If the region is homogeneous, we define a leaf. Otherwise, we have detected a violating cell, and continue splitting until all regions are homogeneous to guarantee faithfulness.
\end{enumerate}
With these adaptations, the heuristic algorithm finds born-again trees that are guaranteed to be faithful but not necessarily minimal in size. Table~\ref{tab:performance-heuristic} compares the computational time of the optimal born-again tree algorithm using objective D ``T$_\text{D}$(s)'', objective L ``T$_\text{L}$(s)'' with that of the heuristic algorithm ``T$_\text{H}$(s)''. It also reports the percentage gap of the heuristic tree depth ``Gap$_\text{D}$(\%)'' and number of leaves ``Gap$_\text{L}$(\%)'' relative to the optimal solution values of each objective.

\begin{table}[htbp]
\caption{Computational time and optimality gap of the heuristic born-again tree algorithm}
\label{tab:performance-heuristic}
\setlength{\tabcolsep}{0.27cm}
\scalebox{0.84}
{
\begin{tabular}{lccccc}
\toprule
Data set & T$_\text{D}$(s) & T$_\text{L}$(s) & T$_\text{H}$(s) & Gap$_\text{D}$(\%) &  Gap$_\text{L}$(\%) \\
\midrule
BC & 39.09 & 1381.45 & 1.14 & 44.80 & 48.37\\
CP & 0.01 & 0.10 & 0.04& 0.00 & 4.85\\
FI & 0.05 & 0.23 & 0.03& 0.00 & 1.79\\
HT & 0.01 & 0.01 & 0.01& 8.33 & 10.92\\
PD & 0.91 & 17.95 & 0.19& 44.79 & 25.63\\
SE & 0.37 & 5.96 & 0.24& 37.25 & 29.03\\
\midrule
Avg. & 6.74 & 234.28& 0.28 & 22.53 & 20.10 \\
\bottomrule
\end{tabular}
}
\end{table}

As visible in these experiments, the CPU time of the heuristic algorithm is significantly smaller than that of the optimal method, at the expense of an increase in tree depth and number of leaves, by 22.53\% and 20.10\% on average, respectively. To test the limits of this heuristic approach, we also verified that it could run in a reasonable amount of time (faster than a minute) on larger datasets such as Ionosphere, Spambase, and Miniboone (the latter with over 130,000 samples and 50 features).
}

\section{Conclusions}
\label{sec:conclusion}

In this paper, we introduced an efficient algorithm to \myblue{transform} a random forest into a single, smallest possible, decision tree. Our algorithm is optimal, and provably returns a single tree with the same \myblue{decision function} as the original tree ensemble. In brief, \myblue{we obtain} a different representation of the \myblue{same} classifier, which helps us to analyze random forests from a different angle. Interestingly, when investigating the structure of the results, we observed that born-again decision trees contain \myblue{many} inexpressive regions designed to faithfully reproduce the behavior of the original ensemble, but which do not contribute to effectively classify samples. It remains an interesting research question to properly understand the purpose of these regions and their contribution to the generalization capabilities of random forests. In a first simple experiment, we attempted to apply post-pruning on the resulting tree. Based on our experiments on six structurally different datasets, we observed that this pruning does not diminish the quality of the predictions but significantly simplifies the born-again trees. Overall, the final pruned trees represent simple, interpretable, and high-performance classifiers, which can be useful for a variety of application areas.

As a perspective for future work, we recommend to \myblue{progress further on solution techniques for the born-again tree ensembles} problem, \myblue{proposing new optimal algorithms} to effectively handle larger datasets \myblue{as well as fast and accurate heuristics. Heuristic upper bounds can also be jointly exploited 
with mathematical programming techniques to eliminate candidate hyperplanes and recursions}. Another interesting research line concerns the combination of the dynamic \myblue{programming} algorithm for the construction of the born-again tree with active pruning during construction, leading to a different definition of the recursion and to different \myblue{base-case evaluations}. Finally, we recommend to pursue the investigation of the structural properties of tree ensembles in light of this new \myblue{born-again tree} representation.

\section*{Acknowledgements}

\myblue{
The authors gratefully thank the editors and referees for their insightful recommendations, as well as Simone Barbosa, Quentin Cappart, and Artur Pessoa for rich scientific discussions. This research has been partially supported by CAPES, CNPq [grant number 308528/2018-2] and FAPERJ [grant number E-26/202.790/2019] in Brazil.
}

\newpage


\onecolumn

\section*{Supplementary Material -- Proofs}
\label{sec:proofs}

\begin{figure}[htbp]
\centering
	\includegraphics[width = 0.9\textwidth]{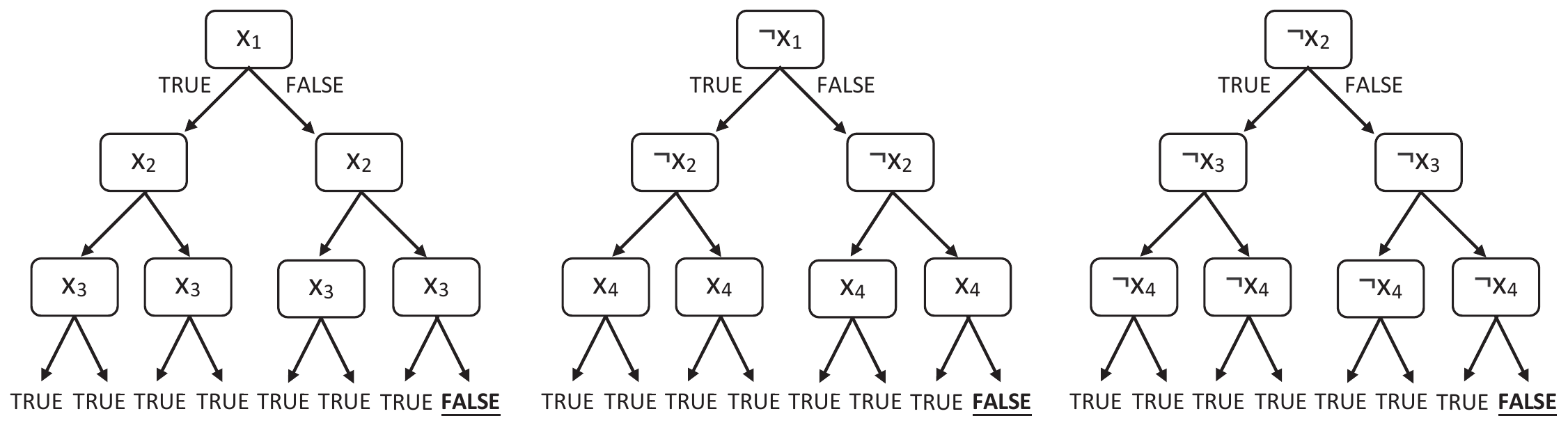}
	\caption{Trees representing the 3-SAT clauses for $\textrm{P} = (x_1 \lor x_2 \lor x_3) \land (\neg x_1 \lor \neg x_2 \lor x_4) \land (x_1 \lor \neg x_3 \lor \neg x_4) $\label{example-clauses}}
\end{figure}

\noindent
\textbf{Proof of Theorem 1.}
We show the NP-hardness of the born-again \myblue{tree ensemble} problem by reduction from \textsc{3-SAT}. Let $\textrm{P}$ be a propositional logic formula presented in conjunctive normal form with three literals per clause. For example, consider $\textrm{P} = (x_1 \lor x_2 \lor x_3) \land (\neg x_1 \lor \neg x_2 \lor x_4) \land (x_1 \lor \neg x_3 \lor \neg x_4)$. 3-SAT aims to determine whether there exist literal values $x_i \in \{\textsc{True},\textsc{False}\}$ in such a way that $\textrm{P}$ is true. From a 3-SAT instance with $k$ clauses and $l$ literals, we construct an instance of the born-again \myblue{tree ensemble} problem with $2k-1$ trees \myblue{of equal weight} as follows:
\begin{itemize}[nosep]
\item As illustrated in Figure~\ref{example-clauses}, the first $k$ trees $(t_1,\dots,t_k)$ represent the clauses. Each of these trees is complete and has a depth of three, with eight leaves representing the possible combinations of values of the three literals. As a consequence of this construction, seven of the leaves predict class \textsc{True}, and the last leaf predicts class \textsc{False}.
\item The last $k-1$ trees contain only a single leaf as root node predicting \textsc{False}.
\end{itemize}

\noindent
Finding the optimal born-again decision tree for this input leads to one of the two following outcomes:
\begin{itemize}[nosep]
\item If the born-again decision tree contains only one leaf predicting class $\textsc{False}$, then 3-SAT for \textrm{P} is $\textsc{False}$.
\item Otherwise 3-SAT for \textrm{P} is $\textsc{True}$.
\end{itemize}

Indeed, in the first case, if the born-again tree only contains a single $\textsc{False}$ region (and since it is faithful to the behavior of the original tree ensemble) there exists no input sample for which $\textsc{True}$ represents the majority class for the $2k-1$ trees. As such, the first $k$ trees cannot jointly predict $\textsc{True}$ for any input and 3-SAT is $\textsc{False}$. 
In the second case, either the optimal born-again decision tree contains a single leaf (root node) of class $\textsc{True}$, or it contains multiple leaves among which at least one leaf predicts $\textsc{True}$ (otherwise the born-again tree would not be optimal). In both situations, there exists a sample for which the majority class of the tree ensemble is $\textsc{True}$ and therefore for which all of the first $k$ trees necessarily return $\textsc{True}$, such that 3-SAT is $\textsc{True}$. This argument holds for any objective involving the minimization of a monotonic size metric, i.e., a metric for which the size of a tree does not decrease upon addition of a node. This includes in particular, the depth, the number of leaves, and the hierarchical objectives involving these two metrics.

\noindent
\textbf{Proof of Theorem 2.}
Consider a tree ensemble \smash{$\mathcal{T} = \{t_1,\dots,t_{|\mathcal{T}|}\}$}. We construct a sequence of decision trees starting with $T_1 = t_1$ by iteratively appending a copy of tree $t_k$ for $k \in \{2,\dots,|\mathcal{T}|\}$ in place of each leaf of the tree $T_{k-1}$ to form~$T_k$. This leads to a born-again tree $T_{|\mathcal{T}|}$ of depth \smash{$\sum_i \Phi(t_i)$}. Each leaf of this tree represents a region of the feature space over which the predicted class of the trees $t \in \mathcal{T}$ is constant, such that the ensemble behavior on this region is faithfully represented by a single class. With this construction, tree $T_{|\mathcal{T}|}$ faithfully reproduces the behavior of the original tree ensemble. Since the optimal born-again tree $T$ has a depth no greater than that of $T_{|\mathcal{T}|}$, we conclude that $\Phi(T) \leq \sum_i \Phi(t_i)$.

Moreover, we prove that this bound is tight, i.e., it is attained for a family of tree ensembles with an arbitrary number of trees. To this end, we consider the feature space $\mathcal{X} = \mathbb{R}^d$ and the following $2d-1$ trees with equal weight:
\begin{itemize}[nosep]
\item For $i \in \{1,\dots,d\}$, tree $t_i$ contains a single internal node representing the split $x_i \leq 0$, leading to a leaf node predicting class $0$ when the splitting condition is satisfied, and to a leaf node predicting class $1$ otherwise.
\item The remaining $d-1$ trees contain a single leaf at the root node predicting class $1$.
\end{itemize}

In the resulting tree ensemble, class $0$ represents the majority if and only if $x_i \leq 0$ for all $i \in \{1,\dots,d\}$. To be faithful to the original tree ensemble, the optimal born-again decision tree must verify that $x_i \leq 0$ for all $i \in \{1,\dots,d\}$ to declare a sample as part of class $0$. This requires at least $d$ comparisons. The depth of the born-again decision tree needed to make these tests is $\Phi(T) = d = \sum_i \Phi(t_i)$.\\

\noindent
\textbf{Proof of Theorem 3.}
Any tree $T$ satisfying $F_T(\mathbf{x}) = F_\mathcal{T}(\mathbf{x})$ on a region $(\mathbf{z}^\textsc{l},\mathbf{z}^\textsc{r})$ also satisfies this condition for any subregion $(\mathbf{\bar{z}}^\textsc{l},\mathbf{\bar{z}}^\textsc{r})$. Therefore, every feasible solution (tree) of the born-again \myblue{tree ensemble} problem for region $(\mathbf{z}^\textsc{l},\mathbf{z}^\textsc{r})$ is feasible for the subproblem restricted to $(\mathbf{\bar{z}}^\textsc{l},\mathbf{\bar{z}}^\textsc{r})$. As a consequence, the optimal solution value $\Phi(\mathbf{\bar{z}}^\textsc{l},\mathbf{\bar{z}}^\textsc{r})$ for the subproblem is smaller or equal than the optimal solution value $\Phi(\mathbf{z}^\textsc{l},\mathbf{z}^\textsc{r})$ of the original problem.\\

\noindent
\textbf{Proof of Theorem 4.} 
We will use the extended notation $\mathds{1}_{jl}(\mathbf{z}^\textsc{l},\mathbf{z}^\textsc{r})$ to denote $\mathds{1}_{jl}$.
Firstly, we observe that $\mathds{1}_{jl}(\mathbf{z}^\textsc{l},\mathbf{z}^\textsc{r}) = \mathds{1}_{jl'}(\mathbf{z}^\textsc{l},\mathbf{z}^\textsc{r})$ for all $l$ and $l'$. Indeed, regardless of $l$ and $j$,
$$
( \Phi(\mathbf{z}^\textsc{l},\mathbf{z}^\textsc{r}_{jl}) =
\Phi(\mathbf{z}^\textsc{l}_{jl},\mathbf{z}^\textsc{r}) = 0 \text{ and } F_\mathcal{T}(\mathbf{z}^\textsc{l}) = F_\mathcal{T}(\mathbf{z}^\textsc{r})) \Leftrightarrow 
\Phi(\mathbf{z}^\textsc{l},\mathbf{z}^\textsc{r}) = 0.
$$

Next, we observe that $\Phi(\mathbf{z}^\textsc{l},\mathbf{z}^\textsc{r}_{jl}) \leq \Phi(\mathbf{z}^\textsc{l},\mathbf{z}^\textsc{r}_{jl'})$ and $\Phi(\mathbf{z}^\textsc{l}_{jl'},\mathbf{z}^\textsc{r}) \leq \Phi(\mathbf{z}^\textsc{l}_{jl},\mathbf{z}^\textsc{r})$ for all $l'>l$ follows from Theorem~3. If $\Phi(\mathbf{z}^\textsc{l},\mathbf{z}^\textsc{r}_{jl}) \geq \Phi(\mathbf{z}^\textsc{l}_{jl},\mathbf{z}^\textsc{r})$, then $\Phi(\mathbf{z}^\textsc{l},\mathbf{z}^\textsc{r}_{jl'}) \geq \Phi(\mathbf{z}^\textsc{l}_{jl'},\mathbf{z}^\textsc{r})$ follows from the two previous inequalities and:
$$
\max \{ 
\Phi(\mathbf{z}^\textsc{l},\mathbf{z}^\textsc{r}_{jl}),
\Phi(\mathbf{z}^\textsc{l}_{jl},\mathbf{z}^\textsc{r})\} = \Phi(\mathbf{z}^\textsc{l},\mathbf{z}^\textsc{r}_{jl}) \leq \Phi(\mathbf{z}^\textsc{l},\mathbf{z}^\textsc{r}_{jl'}) = \max \{ 
\Phi(\mathbf{z}^\textsc{l},\mathbf{z}^\textsc{r}_{jl'}),
\Phi(\mathbf{z}^\textsc{l}_{jl'},\mathbf{z}^\textsc{r})
\}.
$$

Analogously, we observe that based on  Theorem~3 $\Phi(\mathbf{z}^\textsc{l}_{jl},\mathbf{z}^\textsc{r}) \leq \Phi(\mathbf{z}^\textsc{l}_{jl'},\mathbf{z}^\textsc{r})$ and $\Phi(\mathbf{z}^\textsc{l},\mathbf{z}^\textsc{r}_{jl'})\leq \Phi(\mathbf{z}^\textsc{l},\mathbf{z}^\textsc{r}_{jl})$ for all $l'<l$ holds. If $\Phi(\mathbf{z}^\textsc{l},\mathbf{z}^\textsc{r}_{jl}) \leq \Phi(\mathbf{z}^\textsc{l}_{jl},\mathbf{z}^\textsc{r})$, then $\Phi(\mathbf{z}^\textsc{l}_{jl'},\mathbf{z}^\textsc{r}) \geq \Phi(\mathbf{z}^\textsc{l},\mathbf{z}^\textsc{r}_{jl'})$ follows from the two previous inequalities and:
$$
\max \{ 
\Phi(\mathbf{z}^\textsc{l},\mathbf{z}^\textsc{r}_{jl}),
\Phi(\mathbf{z}^\textsc{l}_{jl},\mathbf{z}^\textsc{r})\} = \Phi(\mathbf{z}^\textsc{l}_{jl},\mathbf{z}^\textsc{r}) 
\leq \Phi(\mathbf{z}^\textsc{l}_{jl'},\mathbf{z}^\textsc{r}) = \max \{ 
\Phi(\mathbf{z}^\textsc{l},\mathbf{z}^\textsc{r}_{jl'}),
\Phi(\mathbf{z}^\textsc{l}_{jl'},\mathbf{z}^\textsc{r})
\}.
$$

Combining these results with the first observation, we obtain in both cases that:
 $$\mathds{1}_{jl}(\mathbf{z}^\textsc{l},\mathbf{z}^\textsc{r}) + \max \{ 
\Phi(\mathbf{z}^\textsc{l},\mathbf{z}^\textsc{r}_{jl}),
\Phi(\mathbf{z}^\textsc{l}_{jl},\mathbf{z}^\textsc{r})\} \leq \mathds{1}_{jl'}(\mathbf{z}^\textsc{l},\mathbf{z}^\textsc{r}) + \max \{ 
\Phi(\mathbf{z}^\textsc{l},\mathbf{z}^\textsc{r}_{jl'}),
\Phi(\mathbf{z}^\textsc{l}_{jl'},\mathbf{z}^\textsc{r})
\}.$$

\section*{Supplementary Material -- Pseudo-Codes for Objectives D and DL}
\label{sec:algorithms-results}

Our solution approach is applicable to different tree size metrics, though the binary search argument resulting from Theorem~4 is only applicable for depth minimization. We considered three possible objectives.
\begin{itemize}[nosep]
\item (D) Depth minimization;
\item (L) Minimization of the number of leaves;
\item (DL) Depth minimization \myblue{as primary objective}, and then number of leaves as a secondary objective.
\end{itemize}
The dynamic programming algorithm for depth minimization (D) is described in the main body of the paper. Algorithms~1~and~2 detail the implementation of the dynamic programming algorithm for objectives L and DL, respectively. To maintain a similar structure and use the same solution extraction procedure in Section~\ref{sec:algorithms-extraction}, these two algorithms focus on minimizing the number of splits rather than the number of leaves, given that these quantities are proportional and only differ by one unit \myblue{in a proper binary tree}. Moreover, the hierarchical objective DL is transformed into a weighted sum by associating a large cost of $\textrm{M}$ for each depth increment, and $1$ for each split. This allows to store each dynamic programming result as a single value and reduces memory usage.

\begin{figure*}[htbp]
\begin{minipage}{0.48\textwidth}
\begin{algorithm}[H]
\caption{\textsc{Born-Again-L}($\mathbf{z}^\textsc{l},\mathbf{z}^\textsc{r})$}
\label{algo-DP-L}
\begin{algorithmic}[1]
\STATE \textbf{if} $(\mathbf{z}^\textsc{l} = \mathbf{z}^\textsc{r})$ \textbf{return} 0
\IF{$(\mathbf{z}^\textsc{l},\mathbf{z}^\textsc{r})$ exists in memory}
\STATE \textbf{return} $\textsc{Memory}(\mathbf{z}^\textsc{l},\mathbf{z}^\textsc{r})$
\ENDIF
\STATE $\textsc{UB} \gets \infty$
\STATE $\textsc{LB} \gets 0$
\FOR{$j=1$ {\bfseries to} $p$ \text{ and } $\textsc{LB} < \textsc{UB}$}
	\FOR{$l=z_j^\textsc{l}$ {\bfseries to} $z_j^\textsc{r}-1$ \text{ and } $\textsc{LB} < \textsc{UB}$}
		\STATE $\Phi_1 \gets \textsc{Born-Again-L}(\mathbf{z}^\textsc{l},\mathbf{z}^\textsc{r}+\mathbf{e}_j(l - z^\textsc{r}_j))$
		\STATE $\Phi_2 \gets \textsc{Born-Again-L}(\mathbf{z}^\textsc{l}+\mathbf{e}_j(l + 1 - z^\textsc{l}_j),\mathbf{z}^\textsc{r})$
		\IF {$(\Phi_1 = 0)$ \text{and} $(\Phi_2 = 0)$}
		\STATE \textbf{if} $f(\mathbf{z}^\textsc{l},\mathcal{T}) = f(\mathbf{z}^\textsc{r},\mathcal{T})$ \textbf{then} 
		\STATE \hspace*{0.5cm} \textsc{Memorize}(($\mathbf{z}^\textsc{l},\mathbf{z}^\textsc{r}),0$) \textbf{and return} 0
        \STATE \myblue{\textbf{else}}
		\STATE \hspace*{0.5cm} \textsc{Memorize}(($\mathbf{z}^\textsc{l},\mathbf{z}^\textsc{r}),1$) \textbf{and return} 1
		\STATE \textbf{end if}
		\ENDIF
\STATE 
    \STATE
	\STATE $\textsc{UB} \gets \min \{\textsc{UB},1+\Phi_1+\Phi_2\}$ 
	\STATE $\textsc{LB} \gets \max \{\textsc{LB},\max\{\Phi_1,\Phi_2\}\}$ 
	\ENDFOR
\ENDFOR
	\STATE \textsc{Memorize}(($\mathbf{z}^\textsc{l},\mathbf{z}^\textsc{r}),\textsc{UB}$) \textbf{and return} UB
\end{algorithmic}
\end{algorithm}
\end{minipage}
\hfill
\begin{minipage}{0.5\textwidth}
\begin{algorithm}[H]
\caption{\textsc{Born-Again-DL}($\mathbf{z}^\textsc{l},\mathbf{z}^\textsc{r})$}
\label{algo-DP-DL}
\begin{algorithmic}[1]
\STATE \textbf{if} $(\mathbf{z}^\textsc{l} = \mathbf{z}^\textsc{r})$ \textbf{return} 0
\IF{$(\mathbf{z}^\textsc{l},\mathbf{z}^\textsc{r})$ exists in memory}
\STATE \textbf{return} $\textsc{Memory}(\mathbf{z}^\textsc{l},\mathbf{z}^\textsc{r})$
\ENDIF
\STATE $\textsc{UB} \gets \infty$
\STATE $\textsc{LB} \gets 0$
\FOR{$j=1$ {\bfseries to} $p$ \text{ and } $\textsc{LB} < \textsc{UB}$}
	\FOR{$l=z_j^\textsc{l}$ {\bfseries to} $z_j^\textsc{r}-1$ \text{ and } $\textsc{LB} < \textsc{UB}$}
		\STATE $\Phi_1 \gets \textsc{Born-Again-DL}(\mathbf{z}^\textsc{l},\mathbf{z}^\textsc{r}+\mathbf{e}_j(l - z^\textsc{r}_j))$
		\STATE $\Phi_2 \gets \textsc{Born-Again-DL}(\mathbf{z}^\textsc{l}+\mathbf{e}_j(l + 1 - z^\textsc{l}_j),\mathbf{z}^\textsc{r})$
		\IF {$(\Phi_1 = 0)$ \text{and} $(\Phi_2 = 0)$}
		\STATE \textbf{if} $f(\mathbf{z}^\textsc{l},\mathcal{T}) = f(\mathbf{z}^\textsc{r},\mathcal{T})$ \textbf{then} 
		\STATE \hspace*{0.15cm} \textsc{Memorize}(($\mathbf{z}^\textsc{l},\mathbf{z}^\textsc{r}),0$) \textbf{and return} 0
        \STATE \myblue{\textbf{else}}
		\STATE \hspace*{0.15cm} \textsc{Memorize}(($\mathbf{z}^\textsc{l},\mathbf{z}^\textsc{r}),$\textrm{M}+1) \mbox{\textbf{and return} \textrm{M}+1}
		\STATE \textbf{end if}
		\ENDIF
    \STATE $\textsc{Depth} \gets 1 + \max\{\lfloor \Phi_1/\textrm{M} \rfloor,\lfloor\Phi_2/\textrm{M}\rfloor\}$
    \STATE $\textsc{Splits} \gets 1 + \Phi_1 \% \textrm{M} + \Phi_2 \% \textrm{M}$
	\STATE $\textsc{UB} \gets \min \{\textsc{UB},\textrm{M}\times\textsc{Depth} + \textsc{Splits}\}$ 
	\STATE $\textsc{LB} \gets \max \{\textsc{LB},\max\{\Phi_1,\Phi_2\}\}$ 
	\ENDFOR
\ENDFOR
	\STATE \textsc{Memorize}(($\mathbf{z}^\textsc{l},\mathbf{z}^\textsc{r}),\textsc{UB}$) \textbf{and return} UB
\end{algorithmic}
\end{algorithm}
\end{minipage}
\end{figure*}

The main differences with the algorithm for objective~D occur in the loop of Line~8, which consists for L and DL in an enumeration instead of a binary search. The objective calculations are also naturally different. As seen in Line~20, the new number of splits is calculated as $1+\Phi_1+\Phi_2$ for objective L (i.e., \myblue{the} sum of the splits from the subtrees plus one). When using the hierarchical objective DL, we \myblue{obtain} the depth and number of splits from the subproblems as $\lfloor \Phi_i/\textrm{M} \rfloor$ and $\Phi_i \% \textrm{M}$, respectively, and use these values to obtain the new objective. Our implementation of these algorithms (in C++) is available at the following address: \url{https://github.com/vidalt/BA-Trees}.

\section*{Supplementary Material -- Solution Extraction}
\label{sec:algorithms-extraction}

To reduce computational time and memory consumption, our dynamic programming algorithms only store the optimal objective value of the subproblems. To extract the complete solution, we exploit the following conditions to recursively retrieve the optimal splits from the DP states. For a split to belong to the optimal solution:
\begin{enumerate}[nosep]
\item Both subproblems should exist in the dynamic programming memory.
\item The objective value calculated from the subproblems should match the known optimal value for the considered region.
\end{enumerate}
These conditions lead to Algorithm~3, which reports the optimal tree in DFS order.\vspace*{-0.1cm}

\begin{figure*}[htbp]
\centering
\begin{minipage}{0.68\textwidth}
\begin{algorithm}[H]
\caption{\textsc{Extract-Optimal-Solution}($\mathbf{z}^\textsc{l},\mathbf{z}^\textsc{r},\Phi_\textsc{Opt})$}
\label{algo-DP-L}
\begin{algorithmic}[1]
\IF{$\Phi_\textsc{Opt} = 0$}
\STATE \textsc{Export} a leaf with class $\textsc{Majority-Class}(\mathbf{z}^\textsc{l})$
\STATE \textbf{return}
\ELSE
\FOR{$j=1$ {\bfseries to} $p$}
	\FOR{$l=z_j^\textsc{l}$ {\bfseries to} $z_j^\textsc{r}-1$}
		\STATE $\Phi_1 \gets \textsc{Memory}(\mathbf{z}^\textsc{l},\mathbf{z}^\textsc{r}+\mathbf{e}_j(l - z^\textsc{r}_j))$
		\STATE $\Phi_2 \gets \textsc{Memory}(\mathbf{z}^\textsc{l}+\mathbf{e}_j(l + 1 - z^\textsc{l}_j,\mathbf{z}^\textsc{r})$
		\IF{$\Phi_\textsc{Opt} = \textsc{Calculate-Objective}(\Phi_1,\Phi_2)$}
		\STATE $\textsc{Extract-Optimal-Solution}(\mathbf{z}^\textsc{l},\mathbf{z}^\textsc{r}+\mathbf{e}_j(l - z^\textsc{r}_j),\Phi_1)$
		\STATE $\textsc{Extract-Optimal-Solution}(\mathbf{z}^\textsc{l}+\mathbf{e}_j(l + 1 - z^\textsc{l}_j),\mathbf{z}^\textsc{r},\Phi_2)$
		\STATE \textsc{Export} a split on feature $j$ with level $z_j^\textsc{l}$
		\STATE \textbf{return}		
\ENDIF
	\ENDFOR
\ENDFOR
\ENDIF
\end{algorithmic}
\end{algorithm}
\end{minipage}
\end{figure*}

\section*{Supplementary Material -- Detailed Results}
\label{sec:detailed-results}

In this section, we report additional computational results which did not fit in the main paper due to space limitations. Tables~\ref{tab:EC3}~to~\ref{tab:EC5} extend the results of Tables~2~and~3 in the main paper. They report for each objective the depth and number of leaves of the different classifiers, as well as their minimum and maximum values achieved over the ten runs (one for each training/test pair).\vspace*{-0.3cm}

\begin{table}[htbp]
\caption{Complexity of the different classifiers -- Considering objective D}
\label{tab:EC3}
\setlength{\tabcolsep}{0.15cm}
\centering
\scalebox{0.9}
{
\begin{tabular}{l@{\hspace*{0.6cm}}lll@{\hspace*{0.8cm}}lll@{\hspace*{0.8cm}}lll@{\hspace*{0.8cm}}lll@{\hspace*{0.8cm}}lll}
\toprule
& \multicolumn{3}{@{}l}{Random Forest} & \multicolumn{6}{@{}l}{Born Again Tree} & \multicolumn{6}{@{}l}{Born Again Tree + Pruning} \\
& \multicolumn{3}{@{}l}{\#Leaves}  & \multicolumn{3}{@{}l}{Depth}  &  \multicolumn{3}{@{}l}{\#Leaves} & \multicolumn{3}{@{}l}{Depth}  &  \multicolumn{3}{@{}l}{\#Leaves} \\
Data set & Avg. & Min & Max & Avg. & Min & Max & Avg. & Min & Max & Avg. & Min & Max & Avg. & Min & Max \\
\midrule
BC&61.1&57&68&12.5&11&13&2279.4&541&4091&9.1&8&11&35.9&26&44\\
CP&46.7&40&55&8.9&7&11&119.9&23&347&7.0&4&9&31.2&10&50\\
FI&47.3&40&52&8.6&3&13&71.3&5&269&6.5&3&9&15.8&4&27\\
HT&42.6&36&49&6.0&2&7&20.2&3&38&5.1&2&6&13.2&3&22\\
PD&53.7&45&63&9.6&7&12&460.1&101&1688&9.4&7&12&79.0&53&143\\
SE&55.7&51&60&10.2&9&11&450.9&159&793&7.5&6&8&21.5&16&31\\
\midrule
Overall&51.2&36&68&9.3&2&13&567.0&3&4091&7.4&2&12&32.8&3&143\\
\bottomrule
\end{tabular}%
}
\end{table}%

\begin{table}[htbp]
\caption{Complexity of the different classifiers -- Considering objective L}
\label{tab:EC4}
\setlength{\tabcolsep}{0.15cm}
\centering
\scalebox{0.9}
{
\begin{tabular}{l@{\hspace*{0.6cm}}lll@{\hspace*{0.8cm}}lll@{\hspace*{0.8cm}}lll@{\hspace*{0.8cm}}lll@{\hspace*{0.8cm}}lll}
\toprule
& \multicolumn{3}{@{}l}{Random Forest} & \multicolumn{6}{@{}l}{Born Again Tree} & \multicolumn{6}{@{}l}{Born Again Tree + Pruning} \\
& \multicolumn{3}{@{}l}{\#Leaves}  & \multicolumn{3}{@{}l}{Depth}  &  \multicolumn{3}{@{}l}{\#Leaves} & \multicolumn{3}{@{}l}{Depth}  &  \multicolumn{3}{@{}l}{\#Leaves} \\
Data set & Avg. & Min & Max & Avg. & Min & Max & Avg. & Min & Max & Avg. & Min & Max & Avg. & Min & Max \\
\midrule
BC&61.1&57&68&18.0&17&20&890.1&321&1717&9.0&7&11&23.1&17&32\\
CP&46.7&40&55&8.9&7&11&37.1&10&105&6.5&3&8&11.4&4&21\\
FI&47.3&40&52&8.6&3&13&39.2&4&107&6.3&3&8&12.0&4&20\\
HT&42.6&36&49&6.3&2&8&11.9&3&19&4.3&2&6&6.4&3&9\\
PD&53.7&45&63&15.0&12&19&169.7&50&345&11.0&8&17&30.7&20&42\\
SE&55.7&51&60&13.8&12&16&214.6&60&361&7.7&6&9&14.2&9&19\\
\midrule
Overall&51.2&36&68&11.8&2&20&227.1&3&1717&7.5&2&17&16.3&3&42\\
\bottomrule
\end{tabular}%
}
\end{table}%

\begin{table}[htbp]
\caption{Complexity of the different classifiers -- Considering objective DL}
\label{tab:EC5}
\setlength{\tabcolsep}{0.15cm}
\centering
\scalebox{0.9}
{
\begin{tabular}{l@{\hspace*{0.6cm}}lll@{\hspace*{0.8cm}}lll@{\hspace*{0.8cm}}lll@{\hspace*{0.8cm}}lll@{\hspace*{0.8cm}}lll}
\toprule
& \multicolumn{3}{@{}l}{Random Forest} & \multicolumn{6}{@{}l}{Born Again Tree} & \multicolumn{6}{@{}l}{Born Again Tree + Pruning} \\
& \multicolumn{3}{@{}l}{\#Leaves}  & \multicolumn{3}{@{}l}{Depth}  &  \multicolumn{3}{@{}l}{\#Leaves} & \multicolumn{3}{@{}l}{Depth}  &  \multicolumn{3}{@{}l}{\#Leaves} \\
Data set & Avg. & Min & Max & Avg. & Min & Max & Avg. & Min & Max & Avg. & Min & Max & Avg. & Min & Max \\
\midrule
BC&61.1&57&68&12.5&11&13&1042.3&386&2067&8.9&8&10&27.7&18&39\\
CP&46.7&40&55&8.9&7&11&37.1&10&105&6.5&3&8&11.4&4&21\\
FI&47.3&40&52&8.6&3&13&39.2&4&107&6.3&3&8&12.0&4&20\\
HT&42.6&36&49&6.0&2&7&12.0&3&19&4.6&2&6&6.5&3&10\\
PD&53.7&45&63&9.6&7&12&206.7&70&387&8.9&7&11&42.1&28&62\\
SE&55.7&51&60&10.2&9&11&261.0&65&495&7.4&6&9&17.0&12&24\\
\midrule
Overall&51.2&36&68&9.3&2&13&266.4&3&2067&7.1&2&11&19.5&3&62\\
\bottomrule
\end{tabular}%
}
\end{table}%

Finally, Tables \ref{tab:EC6}~to~\ref{tab:EC8} extend the results of Table~4 in the main paper. They report for each objective the average accuracy and F1 scores of the different classifiers, as well as the associated standard deviations over the ten runs on different training/test pairs.

\begin{table}[htbp]
\caption{Accuracy of the different classifiers -- Considering objective D}
\label{tab:EC6}
\setlength{\tabcolsep}{0.15cm}
\centering
\scalebox{0.9}
{
\begin{tabular}{l@{\hspace*{1cm}}ll@{\hspace*{0.6cm}}ll@{\hspace*{1cm}}ll@{\hspace*{0.6cm}}ll@{\hspace*{1cm}}ll@{\hspace*{0.6cm}}ll}
\toprule
& \multicolumn{4}{@{}l}{Random Forest} & \multicolumn{4}{@{}l}{Born Again Tree} & \multicolumn{4}{@{}l}{Born Again Tree + Pruning} \\
& \multicolumn{2}{@{}l}{Acc.}  &  \multicolumn{2}{@{}l}{F1}  & \multicolumn{2}{@{}l}{Acc.}  &  \multicolumn{2}{@{}l}{F1}  & \multicolumn{2}{@{}l}{Acc.}  &  \multicolumn{2}{@{}l}{F1} \\
Data set & Avg. & Std. & Avg. & Std. &Avg. & Std. &Avg. & Std. &Avg. & Std. &Avg. & Std.\\
\midrule
BC&0.953&0.040&0.949&0.040&0.953&0.040&0.949&0.040&0.946&0.047&0.941&0.046\\
CP&0.660&0.022&0.650&0.024&0.660&0.022&0.650&0.024&0.660&0.022&0.650&0.024\\
FI&0.697&0.049&0.690&0.049&0.697&0.049&0.690&0.049&0.697&0.049&0.690&0.049\\
HT&0.977&0.009&0.909&0.044&0.977&0.009&0.909&0.044&0.977&0.009&0.909&0.044\\
PD&0.746&0.062&0.692&0.065&0.746&0.062&0.692&0.065&0.750&0.067&0.700&0.069\\
SE&0.790&0.201&0.479&0.207&0.790&0.201&0.479&0.207&0.790&0.196&0.481&0.208\\
\midrule
Avg.&0.804&0.064&0.728&0.072&0.804&0.064&0.728&0.072&0.803&0.065&0.729&0.073\\
\bottomrule
\end{tabular}%
}
\end{table}%

\begin{table}[htbp]
\caption{Accuracy of the different classifiers -- Considering objective L}
\label{tab:EC7}
\setlength{\tabcolsep}{0.15cm}
\centering
\scalebox{0.9}
{
\begin{tabular}{l@{\hspace*{1cm}}ll@{\hspace*{0.6cm}}ll@{\hspace*{1cm}}ll@{\hspace*{0.6cm}}ll@{\hspace*{1cm}}ll@{\hspace*{0.6cm}}ll}
\toprule
& \multicolumn{4}{@{}l}{Random Forest} & \multicolumn{4}{@{}l}{Born Again Tree} & \multicolumn{4}{@{}l}{Born Again Tree + Pruning} \\
& \multicolumn{2}{@{}l}{Acc.}  &  \multicolumn{2}{@{}l}{F1}  & \multicolumn{2}{@{}l}{Acc.}  &  \multicolumn{2}{@{}l}{F1}  & \multicolumn{2}{@{}l}{Acc.}  &  \multicolumn{2}{@{}l}{F1} \\
Data set & Avg. & Std. & Avg. & Std. &Avg. & Std. &Avg. & Std. &Avg. & Std. &Avg. & Std.\\
\midrule
BC&0.953&0.040&0.949&0.040&0.953&0.040&0.949&0.040&0.943&0.052&0.938&0.053\\
CP&0.660&0.022&0.650&0.024&0.660&0.022&0.650&0.024&0.660&0.022&0.650&0.024\\
FI&0.697&0.049&0.690&0.049&0.697&0.049&0.690&0.049&0.697&0.049&0.690&0.049\\
HT&0.977&0.009&0.909&0.044&0.977&0.009&0.909&0.044&0.977&0.009&0.909&0.044\\
PD&0.746&0.062&0.692&0.065&0.746&0.062&0.692&0.065&0.751&0.064&0.698&0.068\\
SE&0.790&0.201&0.479&0.207&0.790&0.201&0.479&0.207&0.790&0.193&0.479&0.207\\
\midrule
Avg.&0.804&0.064&0.728&0.072&0.804&0.064&0.728&0.072&0.803&0.065&0.727&0.074\\
\bottomrule
\end{tabular}%
}
\end{table}%

\begin{table}[htbp]
\caption{Accuracy of the different classifiers -- Considering objective DL}
\label{tab:EC8}
\setlength{\tabcolsep}{0.15cm}
\centering
\scalebox{0.9}
{
\begin{tabular}{l@{\hspace*{1cm}}ll@{\hspace*{0.6cm}}ll@{\hspace*{1cm}}ll@{\hspace*{0.6cm}}ll@{\hspace*{1cm}}ll@{\hspace*{0.6cm}}ll}
\toprule
& \multicolumn{4}{@{}l}{Random Forest} & \multicolumn{4}{@{}l}{Born Again Tree} & \multicolumn{4}{@{}l}{Born Again Tree + Pruning} \\
& \multicolumn{2}{@{}l}{Acc.}  &  \multicolumn{2}{@{}l}{F1}  & \multicolumn{2}{@{}l}{Acc.}  &  \multicolumn{2}{@{}l}{F1}  & \multicolumn{2}{@{}l}{Acc.}  &  \multicolumn{2}{@{}l}{F1} \\
Data set & Avg. & Std. & Avg. & Std. &Avg. & Std. &Avg. & Std. &Avg. & Std. &Avg. & Std.\\
\midrule
BC&0.953&0.040&0.949&0.040&0.953&0.040&0.949&0.040&0.941&0.051&0.935&0.049\\
CP&0.660&0.022&0.650&0.024&0.660&0.022&0.650&0.024&0.660&0.022&0.650&0.024\\
FI&0.697&0.049&0.690&0.049&0.697&0.049&0.690&0.049&0.697&0.049&0.690&0.049\\
HT&0.977&0.009&0.909&0.044&0.977&0.009&0.909&0.044&0.977&0.009&0.909&0.044\\
PD&0.746&0.062&0.692&0.065&0.746&0.062&0.692&0.065&0.747&0.069&0.693&0.076\\
SE&0.790&0.201&0.479&0.207&0.790&0.201&0.479&0.207&0.781&0.195&0.477&0.210\\
\midrule
Avg.&0.804&0.064&0.728&0.072&0.804&0.064&0.728&0.072&0.801&0.066&0.726&0.075\\
\bottomrule
\end{tabular}%
}
\end{table}%

\newpage

\section*{Supplementary Material -- Born-Again Tree Illustration}
\label{sec:illustration}

\begin{figure}[htbp]
\centering
	\includegraphics[width = 0.95 \textwidth]{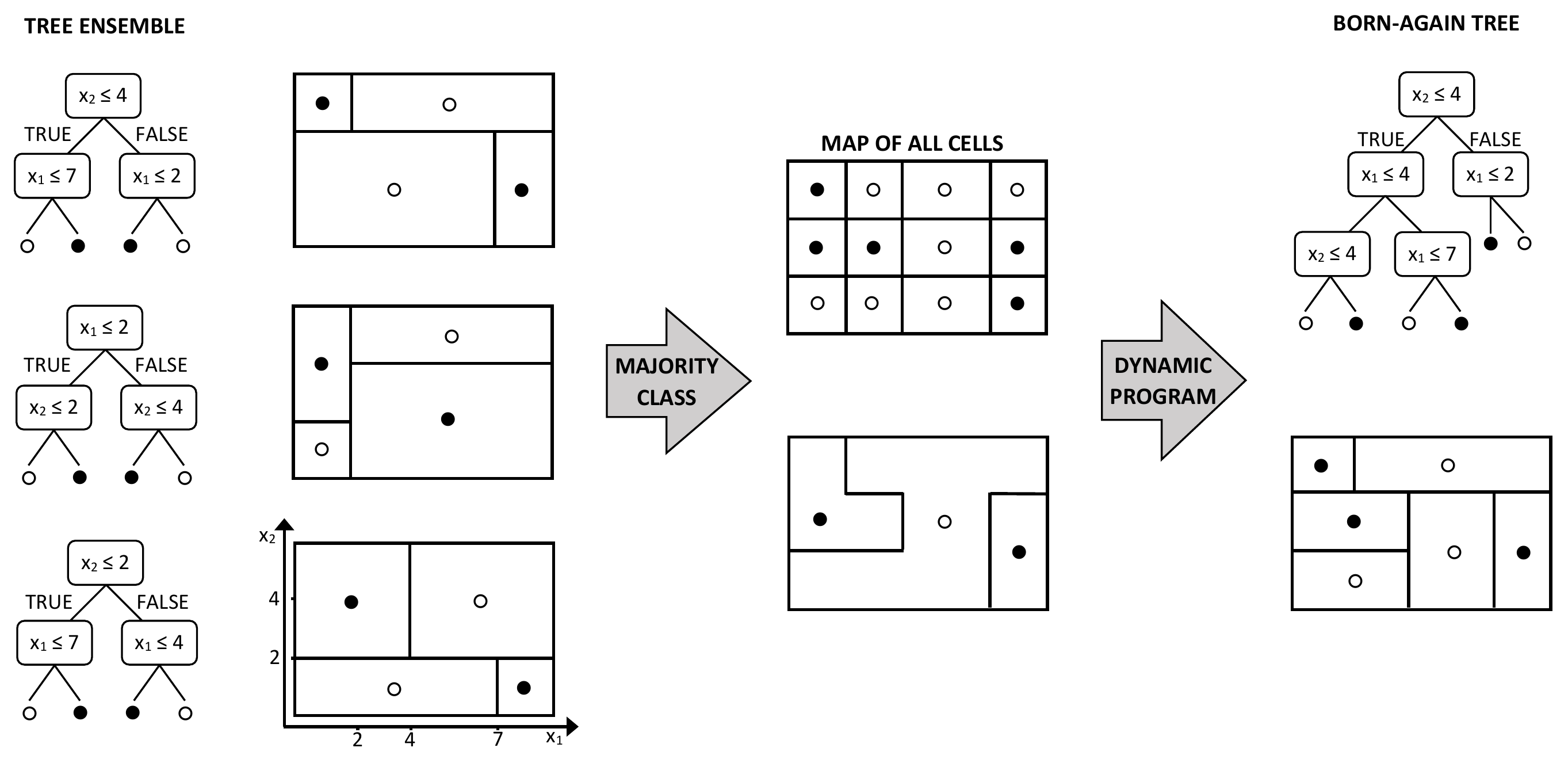}
	\caption{Cells and born-again tree on a simple example \label{fig::illustration}}
\end{figure}

Finally, Figure~\ref{fig::illustration} illustrates the born-again tree ensemble problem on a simple example with an original ensemble composed of three trees. All cells and the corresponding majority classes are represented. There are two classes, depicted by a $\bullet$ and a $\circ$ sign, respectively.

\begin{thebibliography}{53}
\providecommand{\natexlab}[1]{#1}
\providecommand{\url}[1]{\texttt{#1}}
\expandafter\ifx\csname urlstyle\endcsname\relax
  \providecommand{\doi}[1]{doi: #1}\else
  \providecommand{\doi}{doi: \begingroup \urlstyle{rm}\Url}\fi

\bibitem[Bai et~al.(2019)Bai, Li, Li, Jiang, and Xia]{BaiLiEtAl2019}
Bai, J., Li, Y., Li, J., Jiang, Y., and Xia, S.
\newblock Rectified decision trees: Towards interpretability, compression and
  empirical soundness.
\newblock \emph{arXiv preprint arXiv:1903.05965}, 2019.

\bibitem[Banfield et~al.(2005)Banfield, Hall, Bowyer, and
  Kegelmeyer]{BanfieldHallEtAl2005}
Banfield, R.~E., Hall, L.~O., Bowyer, K.~W., and Kegelmeyer, W.~P.
\newblock Ensemble diversity measures and their application to thinning.
\newblock \emph{Information Fusion}, 6\penalty0 (1):\penalty0 49--62, 2005.

\bibitem[Bastani et~al.(2017{\natexlab{a}})Bastani, Kim, and
  Bastani]{BastaniKimEtAl2017}
Bastani, O., Kim, C., and Bastani, H.
\newblock Interpretability via model extraction.
\newblock \emph{arXiv preprint arXiv:1706.09773}, 2017{\natexlab{a}}.

\bibitem[Bastani et~al.(2017{\natexlab{b}})Bastani, Kim, and
  Bastani]{BastaniKimEtAl2017b}
Bastani, O., Kim, C., and Bastani, H.
\newblock Interpreting blackbox models via model extraction.
\newblock \emph{arXiv preprint arXiv:1705.08504}, 2017{\natexlab{b}}.

\bibitem[Bennett(1992)]{Bennett1992}
Bennett, K.
\newblock Decision tree construction via linear programming.
\newblock In \emph{Proceedings of the 4th {M}idwest {Artificial} Intelligence
  and Cognitive Science Society Conference, Utica, Illinois}, 1992.

\bibitem[Bennett \& Blue(1996)Bennett and Blue]{BennettBlue1996}
Bennett, K. and Blue, J.
\newblock Optimal decision trees.
\newblock Technical report, Rensselaer Polytechnique Institute, 1996.

\bibitem[Bertsimas \& Dunn(2017)Bertsimas and Dunn]{BertsimasDunn2017}
Bertsimas, D. and Dunn, J.
\newblock Optimal classification trees.
\newblock \emph{Machine Learning}, 106\penalty0 (7):\penalty0 1039--1082, 2017.

\bibitem[Breiman(2001)]{Breiman2001}
Breiman, L.
\newblock Random forests.
\newblock \emph{Machine Learning}, 45\penalty0 (1):\penalty0 5--32, 2001.

\bibitem[Breiman \& Shang(1996)Breiman and Shang]{BreimanShang1996}
Breiman, L. and Shang, N.
\newblock Born again trees.
\newblock Technical report, University of California Berkeley, 1996.

\bibitem[Bucilu\v{a} et~al.(2006)Bucilu\v{a}, Caruana, and
  Niculescu-Mizil]{BuciluaCaruanaEtAl2006}
Bucilu\v{a}, C., Caruana, R., and Niculescu-Mizil, A.
\newblock Model compression.
\newblock In \emph{Proceedings of the 12th ACM SIGKDD International Conference
  on Knowledge Discovery and Data Mining}, 2006.

\bibitem[Caruana et~al.(2004)Caruana, Niculescu-Mizil, Crew, and
  Ksikes]{CaruanaNiculescu-MizilEtAl2004}
Caruana, R., Niculescu-Mizil, A., Crew, G., and Ksikes, A.
\newblock Ensemble selection from libraries of models.
\newblock In \emph{Proceedings of the twenty-first International Conference on
  Machine Learning}, pp.\ ~18, 2004.

\bibitem[Clark et~al.(2019)Clark, Luong, Khandelwal, Manning, and
  Le]{ClarkLuongEtAl2019}
Clark, K., Luong, M.-T., Khandelwal, U., Manning, C.~D., and Le, Q.~V.
\newblock Bam! born-again multi-task networks for natural language
  understanding.
\newblock \emph{arXiv preprint arXiv:1907.04829}, 2019.

\bibitem[Frankle \& Carbin(2018)Frankle and Carbin]{FrankleCarbin2018}
Frankle, J. and Carbin, M.
\newblock The lottery ticket hypothesis: Finding sparse, trainable neural
  networks.
\newblock \emph{arXiv preprint arXiv:1803.03635}, 2018.

\bibitem[Friedman(2001)]{Friedman2001}
Friedman, J.
\newblock Greedy function approximation: {A} gradient boosting machine.
\newblock \emph{Annals of Statistics}, 29\penalty0 (5):\penalty0 1189--1232,
  2001.

\bibitem[Friedman \& Popescu(2008)Friedman and Popescu]{FriedmanPopescu2008}
Friedman, J.~H. and Popescu, B.~E.
\newblock Predictive learning via rule ensembles.
\newblock \emph{The Annals of Applied Statistics}, 2\penalty0 (3):\penalty0
  916--954, 2008.

\bibitem[Frosst \& Hinton(2017)Frosst and Hinton]{FrosstHinton2017}
Frosst, N. and Hinton, G.
\newblock Distilling a neural network into a soft decision tree.
\newblock \emph{arXiv preprint arXiv:1711.09784}, 2017.

\bibitem[Furlanello et~al.(2018)Furlanello, Lipton, Tschannen, Itti, and
  Anandkumar]{FurlanelloLiptonEtAl2018}
Furlanello, T., Lipton, Z.~C., Tschannen, M., Itti, L., and Anandkumar, A.
\newblock Born again neural networks.
\newblock \emph{arXiv preprint arXiv:1805.04770}, 2018.

\bibitem[Guidotti et~al.(2018)Guidotti, Monreale, Ruggieri, Turini, Giannotti,
  and Pedreschi]{GuidottiMonrealeEtAl2018}
Guidotti, R., Monreale, A., Ruggieri, S., Turini, F., Giannotti, F., and
  Pedreschi, D.
\newblock A survey of methods for explaining black box models.
\newblock \emph{ACM Computing Surveys (CSUR)}, 51\penalty0 (5):\penalty0 1--42,
  2018.

\bibitem[G{\"u}nl{\"u}k et~al.(2018)G{\"u}nl{\"u}k, Kalagnanam, Menickelly, and
  Scheinberg]{GuenluekKalagnanamEtAl2018}
G{\"u}nl{\"u}k, O., Kalagnanam, J., Menickelly, M., and Scheinberg, K.
\newblock Optimal decision trees for categorical data via integer programming.
\newblock \emph{arXiv preprint arXiv:1612.03225}, 2018.

\bibitem[Hara \& Hayashi(2016)Hara and Hayashi]{HaraHayashi2016}
Hara, S. and Hayashi, K.
\newblock Making tree ensembles interpretable: A bayesian model selection
  approach.
\newblock \emph{arXiv preprint arXiv:1606.09066}, 2016.

\bibitem[Hern\'{a}ndez-Lobato et~al.(2009)Hern\'{a}ndez-Lobato, Martinez-Muoz,
  and Su\'{a}rez]{Hernandez-LobatoMartinez-MuozEtAl2009}
Hern\'{a}ndez-Lobato, D., Martinez-Muoz, G., and Su\'{a}rez, A.
\newblock Statistical instance-based pruning in ensembles of independent
  classifiers.
\newblock \emph{IEEE Transactions on Pattern Analysis and Machine
  Intelligence}, 31\penalty0 (2):\penalty0 364--369, 2009.

\bibitem[Hinton et~al.(2015)Hinton, Vinyals, and Dean]{HintonVinyalsEtAl2015}
Hinton, G., Vinyals, O., and Dean, J.
\newblock Distilling the knowledge in a neural network.
\newblock \emph{arXiv preprint arXiv:1503.02531}, 2015.

\bibitem[Hu et~al.(2007)Hu, Yu, Xie, and Li]{HuYuEtAl2007}
Hu, Q., Yu, D., Xie, Z., and Li, X.
\newblock Eros: Ensemble rough subspaces.
\newblock \emph{Pattern recognition}, 40\penalty0 (12):\penalty0 3728--3739,
  2007.

\bibitem[Hu et~al.(2019)Hu, Rudin, and Seltzer]{HuRudinEtAl2019}
Hu, X., Rudin, C., and Seltzer, M.
\newblock Optimal sparse decision trees.
\newblock In \emph{Advances in Neural Information Processing Systems}, 2019.

\bibitem[Kisamori \& Yamazaki(2019)Kisamori and Yamazaki]{KisamoriYamazaki2019}
Kisamori, K. and Yamazaki, K.
\newblock Model bridging: To interpretable simulation model from neural
  network.
\newblock \emph{arXiv preprint arXiv:1906.09391}, 2019.

\bibitem[Margineantu \& Dietterich(1997)Margineantu and
  Dietterich]{MargineantuDietterich1997}
Margineantu, D. and Dietterich, T.
\newblock Pruning adaptive boosting.
\newblock In \emph{Proceedings of the Fourteenth International Conference
  Machine Learning}, 1997.

\bibitem[Mart{\'\i}nez-Mu{\~n}oz et~al.(2008)Mart{\'\i}nez-Mu{\~n}oz,
  Hern{\'a}ndez-Lobato, and Su{\'a}rez]{Martinez-MunozHernandez-LobatoEtAl2008}
Mart{\'\i}nez-Mu{\~n}oz, G., Hern{\'a}ndez-Lobato, D., and Su{\'a}rez, A.
\newblock An analysis of ensemble pruning techniques based on ordered
  aggregation.
\newblock \emph{IEEE Transactions on Pattern Analysis and Machine
  Intelligence}, 31\penalty0 (2):\penalty0 245--259, 2008.

\bibitem[Meinshausen(2010)]{Meinshausen2010}
Meinshausen, N.
\newblock Node harvest.
\newblock \emph{The Annals of Applied Statistics}, pp.\  2049--2072, 2010.

\bibitem[Melis \& Jaakkola(2018)Melis and Jaakkola]{MelisJaakkola2018}
Melis, D.~A. and Jaakkola, T.
\newblock Towards robust interpretability with self-explaining neural networks.
\newblock In \emph{Advances in Neural Information Processing Systems}, 2018.

\bibitem[Mollas et~al.(2019)Mollas, Tsoumakas, and
  Bassiliades]{MollasTsoumakasEtAl2019}
Mollas, I., Tsoumakas, G., and Bassiliades, N.
\newblock Lionforests: Local interpretation of random forests through path
  selection.
\newblock \emph{arXiv preprint arXiv:1911.08780}, 2019.

\bibitem[Murdoch et~al.(2019)Murdoch, Singh, Kumbier, Abassi-Asl, and
  Yu]{MurdochSinghEtAl2019}
Murdoch, W., Singh, C., Kumbier, K., Abassi-Asl, R., and Yu, B.
\newblock Interpretable machine learning: definitions, methods, and
  applications.
\newblock \emph{arXiv preprint arXiv:1901.04592v1}, 2019.

\bibitem[Nijssen \& Fromont(2007)Nijssen and Fromont]{NijssenFromont2007}
Nijssen, S. and Fromont, E.
\newblock Mining optimal decision trees from itemset lattices.
\newblock In \emph{Proceedings of the 13th ACM SIGKDD International Conference
  on Knowledge Discovery and Data Mining}, 2007.

\bibitem[Park \& Furnkranz(2012)Park and Furnkranz]{ParkFurnkranz2012}
Park, S. and Furnkranz, J.
\newblock Efficient prediction algorithms for binary decomposition techniques.
\newblock \emph{Data Mining and Knowledge Discovery}, 24\penalty0 (1):\penalty0
  40--77, 2012.

\bibitem[Partalas et~al.(2010)Partalas, Tsoumakas, and
  Vlahavas]{PartalasTsoumakasEtAl2010}
Partalas, I., Tsoumakas, G., and Vlahavas, I.
\newblock An ensemble uncertainty aware measure for directed hill climbing
  ensemble pruning.
\newblock \emph{Machine Learning}, 81\penalty0 (3):\penalty0 257--282, 2010.

\bibitem[Prodromidis \& Stolfo(2001)Prodromidis and
  Stolfo]{ProdromidisStolfo2001}
Prodromidis, A.~L. and Stolfo, S.~J.
\newblock Cost complexity-based pruning of ensemble classifiers.
\newblock \emph{Knowledge and Information Systems}, 3\penalty0 (4):\penalty0
  449--469, 2001.

\bibitem[Prodromidis et~al.(1999)Prodromidis, Stolfo, and
  Chan]{ProdromidisStolfoEtAl1999}
Prodromidis, A.~L., Stolfo, S.~J., and Chan, P.~K.
\newblock Effective and efficient pruning of metaclassifiers in a distributed
  data mining system.
\newblock \emph{Knowledge Discovery and Data Mining Journal}, 32, 1999.

\bibitem[Rokach(2009)]{Rokach2009}
Rokach, L.
\newblock Collective-agreement-based pruning of ensembles.
\newblock \emph{Computational Statistics \& Data Analysis}, 53\penalty0
  (4):\penalty0 1015--1026, 2009.

\bibitem[Rokach(2016)]{Rokach2016}
Rokach, L.
\newblock Decision forest: Twenty years of research.
\newblock \emph{Information Fusion}, 27:\penalty0 111--125, 2016.

\bibitem[Rokach et~al.(2006)Rokach, Maimon, and Arbel]{RokachMaimonEtAl2006}
Rokach, L., Maimon, O., and Arbel, R.
\newblock Selective voting—getting more for less in sensor fusion.
\newblock \emph{International Journal of Pattern Recognition and Artificial
  Intelligence}, 20\penalty0 (03):\penalty0 329--350, 2006.

\bibitem[Rudin(2019)]{Rudin2019}
Rudin, C.
\newblock Stop explaining black box machine learning models for high stakes
  decisions and use interpretable models instead.
\newblock \emph{Nature Machine Intelligence}, 1\penalty0 (5):\penalty0
  206--215, 2019.

\bibitem[Sirikulviriya \& Sinthupinyo(2011)Sirikulviriya and
  Sinthupinyo]{SirikulviriyaSinthupinyo2011}
Sirikulviriya, N. and Sinthupinyo, S.
\newblock Integration of rules from a random forest.
\newblock In \emph{International Conference on Information and Electronics
  Engineering}, volume~6, 2011.

\bibitem[Smith et~al.(1988)Smith, Everhart, Dickson, Knowler, and
  Johannes]{Smith1988}
Smith, J., Everhart, J., Dickson, W., Knowler, W., and Johannes, R.
\newblock {Using the ADAP learning algorithm to forecast the onset of diabetes
  mellitus}.
\newblock In \emph{Proceedings of the Annual Symposium on Computer Applications
  in Medical Care}, pp.\  261--265. IEEE Computer Society Press, 1988.

\bibitem[Tamon \& Xiang(2000)Tamon and Xiang]{TamonXiang2000}
Tamon, C. and Xiang, J.
\newblock On the boosting pruning problem.
\newblock In \emph{Proceedings of the 11th European Conference on Machine
  Learning}, 2000.

\bibitem[Tan et~al.(2016)Tan, Hooker, and Wells]{TanHookerEtAl2016}
Tan, H.~F., Hooker, G., and Wells, M.~T.
\newblock Tree space prototypes: Another look at making tree ensembles
  interpretable.
\newblock \emph{arXiv preprint arXiv:1611.07115}, 2016.

\bibitem[Vandewiele et~al.(2017)Vandewiele, Lannoye, Janssens, Ongenae,
  De~Turck, and Van~Hoecke]{VandewieleLannoyeEtAl2017}
Vandewiele, G., Lannoye, K., Janssens, O., Ongenae, F., De~Turck, F., and
  Van~Hoecke, S.
\newblock A genetic algorithm for interpretable model extraction from decision
  tree ensembles.
\newblock In \emph{Pacific-Asia Conference on Knowledge Discovery and Data
  Mining}. Springer, 2017.

\bibitem[Verwer \& Zhang(2019)Verwer and Zhang]{VerwerZhang2019}
Verwer, S. and Zhang, Y.
\newblock Learning optimal classification trees using a binary linear program
  formulation.
\newblock In \emph{Proceedings of the AAAI Conference on Artificial
  Intelligence}, 2019.

\bibitem[Windeatt \& Ardeshir(2001)Windeatt and Ardeshir]{WindeattArdeshir2001}
Windeatt, T. and Ardeshir, G.
\newblock An empirical comparison of pruning methods for ensemble classifiers.
\newblock In \emph{International Symposium on Intelligent Data Analysis}, 2001.

\bibitem[Zhang \& Wang(2009)Zhang and Wang]{ZhangWang2009}
Zhang, H. and Wang, M.
\newblock Search for the smallest random forest.
\newblock \emph{Statistics and its Interface}, 2\penalty0 (3):\penalty0 381,
  2009.

\bibitem[Zhang et~al.(2018)Zhang, Nian~Wu, and Zhu]{ZhangNianWuEtAl2018}
Zhang, Q., Nian~Wu, Y., and Zhu, S.-C.
\newblock Interpretable convolutional neural networks.
\newblock In \emph{Proceedings of the IEEE Conference on Computer Vision and
  Pattern Recognition}, 2018.

\bibitem[Zhang et~al.(2006)Zhang, Burer, and Street]{ZhangBurerEtAl2006}
Zhang, Y., Burer, S., and Street, W.~N.
\newblock Ensemble pruning via semi-definite programming.
\newblock \emph{Journal of Machine Learning Research}, 7\penalty0
  (Jul):\penalty0 1315--1338, 2006.

\bibitem[Zhou \& Hooker(2016)Zhou and Hooker]{ZhouHooker2016}
Zhou, Y. and Hooker, G.
\newblock Interpreting models via single tree approximation.
\newblock \emph{arXiv preprint arXiv:1610.09036}, 2016.

\bibitem[Zhou et~al.(2002)Zhou, Wu, and Tang]{ZhouWuEtAl2002}
Zhou, Z., Wu, J., and Tang, W.
\newblock Ensembling neural networks: many could be better than all.
\newblock \emph{Artificial Intelligence}, 137:\penalty0 239--263, 2002.

\bibitem[Zhou \& Tang(2003)Zhou and Tang]{ZhouTang2003}
Zhou, Z.-H. and Tang, W.
\newblock Selective ensemble of decision trees.
\newblock In \emph{International Workshop on Rough Sets, Fuzzy Sets, Data
  Mining, and Granular-Soft Computing}, 2003.

\end{thebibliography}
\end{document}